\documentclass{article}

\PassOptionsToPackage{numbers, compress}{natbib}
 \usepackage[dblblindworkshop, final]{neurips_2025}
\workshoptitle{Foundations of Reasoning in Language Models}

\usepackage{wrapfig}
\usepackage{dblfloatfix}
\usepackage{xspace}
\usepackage{caption}
\usepackage{subcaption}
\usepackage[utf8]{inputenc} % allow utf-8 input
\usepackage[T1]{fontenc}    % use 8-bit T1 fonts
\usepackage[hidelinks,breaklinks=true,colorlinks,bookmarks=false,citecolor=citecolor,linkcolor=linkcolor]{hyperref}
\usepackage{url}            % simple URL typesetting
\usepackage{booktabs}       % professional-quality tables
\usepackage{amsfonts}       % blackboard math symbols
\usepackage{nicefrac}       % compact symbols for 1/2, etc.
\usepackage{microtype}      % microtypography
\usepackage{xcolor}         % colors
\usepackage{graphicx}
\usepackage{float}
\usepackage{natbib}
\usepackage{verbatim}
\usepackage{newunicodechar}
\newunicodechar{，}{,}
\usepackage{array}
\usepackage{ragged2e}
\usepackage{amsfonts}       
\usepackage{amssymb}        
\usepackage{amsmath}
\usepackage{cleveref}

\usepackage{adjustbox}
\usepackage[most]{tcolorbox}
\usepackage{tikz}           
\usepackage{etoolbox}

\newcolumntype{C}[1]{>{\centering\arraybackslash}m{#1}}
\usepackage{color, colortbl}
\usepackage[shortlabels]{enumitem}
\usepackage{boxedminipage}
\usepackage{multirow}
\usepackage{makecell}

\usepackage{pgfplots}
\pgfplotsset{compat=1.17}
\usepackage[table]{xcolor} % For cell coloring
\usepackage{colortbl}      % For cell coloring
\usepackage{graphicx}      % For adjustbox
\usepackage{adjustbox}     % To fit table to width
\usepackage{booktabs}      % For \toprule, \midrule, \bottomrule

\definecolor{citecolor}{HTML}{2980b9}
\definecolor{linkcolor}{HTML}{c0392b}
\definecolor{darkorange}{HTML}{FF8C00}
\definecolor{chocolate}{HTML}{D2691E}
\definecolor{darkgreen}{HTML}{006400}
\definecolor{darkblue}{HTML}{00008B}
\definecolor{mediumblue}{HTML}{0000CD}
\definecolor{dodgerblue}{HTML}{1E90FF}
\definecolor{royalblue}{HTML}{4169E1}
\definecolor{shadecolor}{RGB}{237,237,237}
\definecolor{backred}{RGB}{255, 190, 190}
\definecolor{backblue}{RGB}{210, 230, 250}
\definecolor{zrrgreen}{HTML}{008000}
\definecolor{zrrblue}{HTML}{4682B4}
\definecolor{zrrred}{HTML}{B22222}
\definecolor{light_green}{HTML}{b2ffb2}
\definecolor{sage}{HTML}{c3efb2}
\definecolor{light_red}{HTML}{ffb2b2}
\definecolor{light_blue}{HTML}{add8e6} 

\definecolor{hgreen}{HTML}{C9F6C8}
\definecolor{hred}{HTML}{FDCFC8}

\tcbset{on line, 
        boxsep=1pt, left=0pt,right=0pt,top=0pt,bottom=0pt,
        colframe=white,
        colback=light_green,  
        highlight math style={enhanced}
        }

\newcommand{\dataset}{\textsc{PolyMATH}\xspace}

\title{\dataset: A Challenging Multi-modal \\ Mathematical Reasoning Benchmark }

\author
{Himanshu Gupta$^{1*}$ \quad Shreyas Verma$^{2*}$ \quad Ujjwala Anantheswaran$^{1*}$ \quad \textbf{Kevin Scaria}$^{1*}$\quad \\ \textbf{Mihir Parmar}$^{1}$ \quad \textbf{Swaroop Mishra}$^{1}$ \quad \textbf{Chitta Baral}$^{1}$  \\
\small{$^{1}$Arizona State University} \quad
\small{$^{2}$Georgia Institute of Technology}\\
\tt\small { \{hgupta35, kscaria\}\@asu.edu }
}

\begin{document}

\maketitle

\begin{abstract}
Multi-modal Large Language Models (MLLMs) exhibit impressive problem-solving abilities in various domains, but their visual comprehension and abstract reasoning skills remain under-evaluated.
To this end, we present \dataset, a challenging benchmark aimed at evaluating the general cognitive reasoning abilities of MLLMs. 
\dataset comprises 5,000 manually collected high-quality images of cognitive textual and visual challenges across 10 distinct categories, including pattern recognition, spatial reasoning, and relative reasoning. 
We conducted a comprehensive, and quantitative evaluation of 12 MLLMs using four diverse prompting strategies, including Chain-of-Thought and Step-Back.
% \ua{edit numbers below}
The best scores achieved on \dataset are $\sim$ 54\%, $\sim$ 36\%, and $\sim$ 57\%, obtained by Claude-3.7 Sonnet, GPT-4o and Gemini-2.5 Flash respectively - highlighting the logical and visual complexity of these questions.
A further fine-grained error analysis reveals that these models struggle to understand spatial relations and perform drawn-out, high-level reasoning.
This is further strengthened by our ablation study estimating MLLM performance when given textual descriptions in place of diagrams.
As evidenced by $\sim$ 4\% improvement over textual descriptions as opposed to actual images, we discover that models do not truly comprehend visual diagrams and the spatial information therein, and are thus prone to logical errors. 
% Finally, we evaluate the OpenAI o1 models and find that their performance only matches the human baseline, highlighting the difficulty of the benchmark.
The results on \dataset highlight the room for improvement in multi-modal reasoning and provide unique insights to guide the development of future MLLMs
\footnote{Codebase: \url{https://github.com/kevinscaria/PolyMATH} \\
Dataset: \url{https://huggingface.co/datasets/him1411/polymath} \\
\quad $*$Equal Contribution 
% \quad $\dagger$ Currently in Google DeepMind 
% \quad $\ddagger$ Currently in Amazon (The work was done prior to joining Amazon) \quad $\diamondsuit$ Currently in Microsoft 
}.

% \footnote{\href{https://anonymous.4open.science/r/PolyMATH-052D}{https://anonymous.4open.science/r/PolyMATH-052D}}.
\end{abstract}

\vspace{-1.2em}  % Adjust this as needed
% {\centering \OpenLink\par}

\begin{figure*}[ht!]
	% \centering
	\includegraphics[width = \linewidth, height= 8.5 cm]{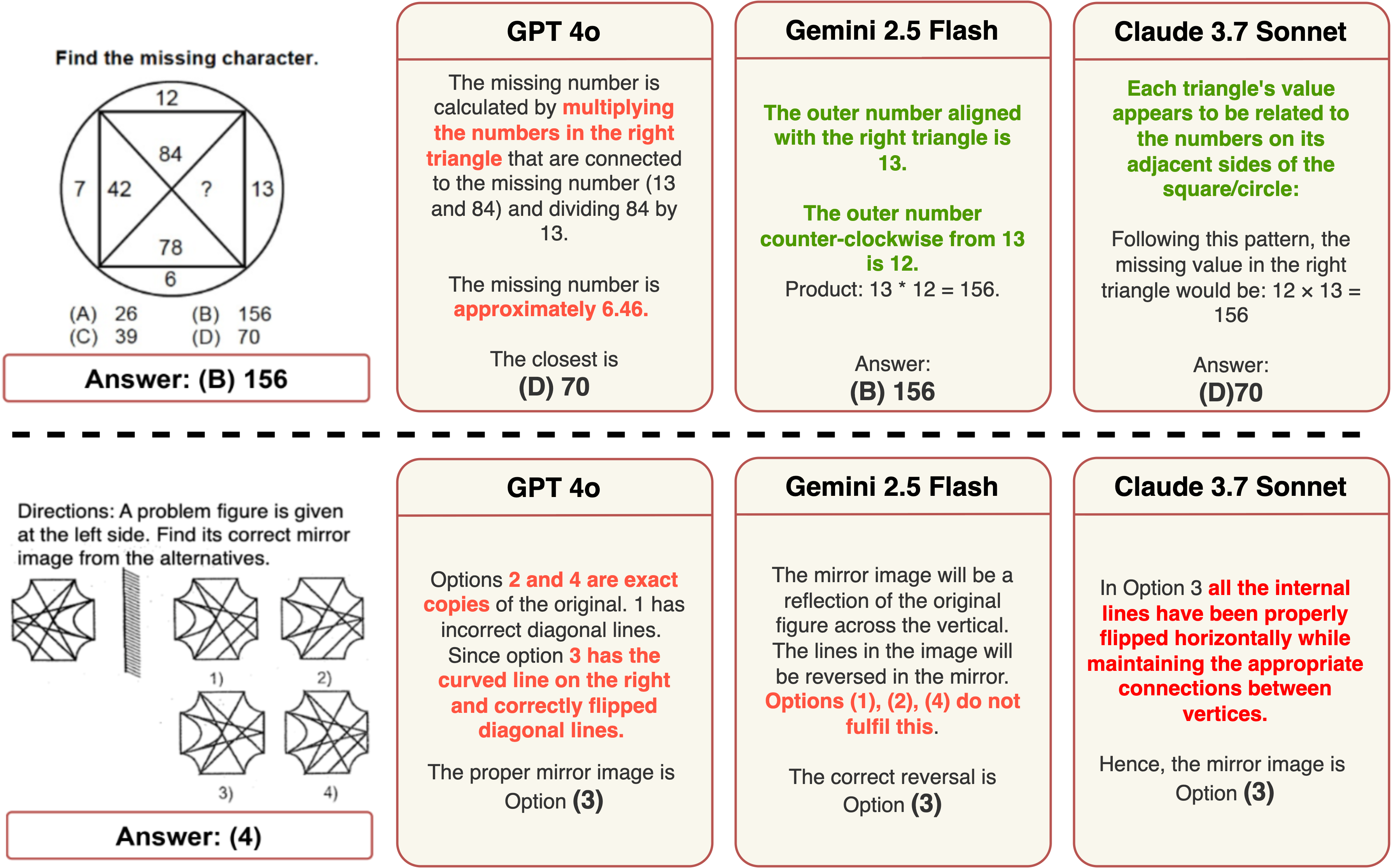}
	\caption{Examples of the reasoning patterns employed by MLLMs when faced with questions involving visual information. In the top row, models fail to perceive the relationship between adjacent semicircles; in the bottom row, models fail to comprehend fine details in the answer images.}
	\label{fig1:teaser}
\end{figure*}

\section{Introduction}

Large Language Models (LLMs) \citep{brown2020language,albert24mixtral,touvron2023llama,achiam2023gpt} and Multi-modal Large Language Models (MLLMs) \citep{openai2023gpt4v,team2023gemini,su2023pandagpt,chen2023minigpt} have rapidly become a pivotal area of research. 
MLLMs with robust reasoning capabilities in visual contexts can solve complex educational problems \citep{seo2015solving,wang2017deep}, support analysts with logical queries on statistical data \citep{wu2023bloomberggpt,yang2023fingpt}, and contribute to advanced research areas such as theorem proving and scientific discovery \citep{taylor2022galactica,dong2023large,trinh2024solving}. 
Despite their impressive performance in various assessments of human-like intelligence, these models still exhibit notable shortcomings on tasks requiring cognitive and logical reasoning, such as commonsense numerical reasoning, scientific problem-solving, and abstract puzzles \citep{wang2023scibench,Lu2023MathVistaEM}. 
Existing evaluation benchmarks \citep{fu2023mme,liu2023mmbench,Li2023SEEDBenchBM,fu2023challenger,sun2024journeydb} have focused primarily on specific concrete domains. 
While general-purpose visual question-answering (VQA) datasets capture some elements of mathematical reasoning, a systematic investigation into abstract and general cognitive reasoning which are essential for tasks like visual puzzles remains an underexplored frontier.

In this paper, we present \dataset, a benchmark specifically crafted to evaluate the complex multi-modal cognitive reasoning capabilities of MLLMs. 
We propose a task taxonomy to guide the development of \dataset: 
(1) we identify ten distinct reasoning skills, including \textit{spatial reasoning}, \textit{pattern recognition}, and \textit{numerical reasoning}.    
and (2) we cover a diverse array of visual contexts, including images with venn diagrams, spatially-related layouts, as well as geometric figures. 
\dataset is a meticulously curated dataset of 5000 multimodal reasoning problems newly acquired from a publicly available source (Table \ref{tab:all_dataset_stats}).
The problems of the original source have been crafted and rigorously reviewed by expert annotators, and require diverse fine-grained problem-solving capabilities.
Additionally, we provide detailed textual representations of diagrams of the samples. 
% This makes \dataset a multi-purpose dataset that is challenging for contemporary models of all modalities. 
As denoted in fig. \ref{fig1:teaser}, these problems are designed to assess the logical reasoning abilities of the average high school student over text and diagrams. 
We observe that MLLMs fail to demonstrate the cognitive reasoning skills required to solve these questions.
% , especially in the context of spatial information.

% We conduct extensive experiments on \dataset with state-of-the-art (SOTA) closed-source MLLMs like the Claude family (3.5 Sonnet, 3 Sonnet, 3 Haiku), Gemini-2.5 Flash, and GPT-4o, and 9 open-source MLLMs like LLaVA (34B) and ShareGPT4V. 
We conduct extensive experiments on \dataset with state-of-the-art (SOTA) closed-source MLLMs like the Claude 3.7 Sonnet, Gemini-2.5 Flash, and GPT-4o, and 9 open-source MLLMs like LLaVA (34B) and ShareGPT4V. 
We evaluate them via zero shot, few shot, Chain-of-Thought \citep{wei2022chain} and step back prompting \citep{zhengtake}.
We show that \dataset is a challenging benchmark, with human performance (established by qualified human annotators with graduate degrees) reaching only 66.3\% accuracy.
The most powerful model we evaluate, Gemini-2.5 Flash, achieves the best score of 57.00\% followed by Claude 3.7 Sonnet, which attains 53.90\%. 
The best open source models like LLaVA-v1.6 Mistral (7B) and ShareGPT4V (13B) achieves the accuracy of 15.20\% and 12.80\% respectively.
We additionally create a diagram only subset (\textit{test-img})
of the benchmark to gauge the gap in visual reasoning abilities between the multi-modal models and average human capability. 
We find that the performance of these models drops further to 26.20\% for Claude-3.7 Sonnet and 32.50\% by Gemini-2.5 Flash when evaluated on \textit{test-img} only.
In contrast with human cognitive patterns, when given text descriptions in place of the diagram in these questions, model accuracy improves by $\sim$4-7\%.
We also conduct an error analysis on Claude-3.7 Sonnet, Gemini-2.5 Flash and GPT-4o, and find that the most common errors stem from misunderstanding diagrams ($\sim 60\%$), misidentifying logical patterns ($\sim 25\%$), and forgetting relational information ($\sim 12\%$).
% Furthermore, these models make identical errors on $\sim 80\%$ of the incorrect pattern recognition questions.
% Finally, we evaluate OpenAI o1 models \citep{openaio1} on without diagram questions of the benchmark and observe 66.72\% accuracy (o1-preview), 2\% points below than the human baseline.
% Additionally, we observe that model performance on diagram based questions [\% of qs here] is [x]\% worse than performance on purely textual questions, exposing the gap in visual reasoning abilities between the most powerful multi-modal models and average human capability.

\begin{figure}[t!]
    \vspace{-4mm}
        \begin{minipage}[t]{0.49\linewidth}
        \centering
        \includegraphics[
        width = 6 cm, height= 6 cm
        ]
        {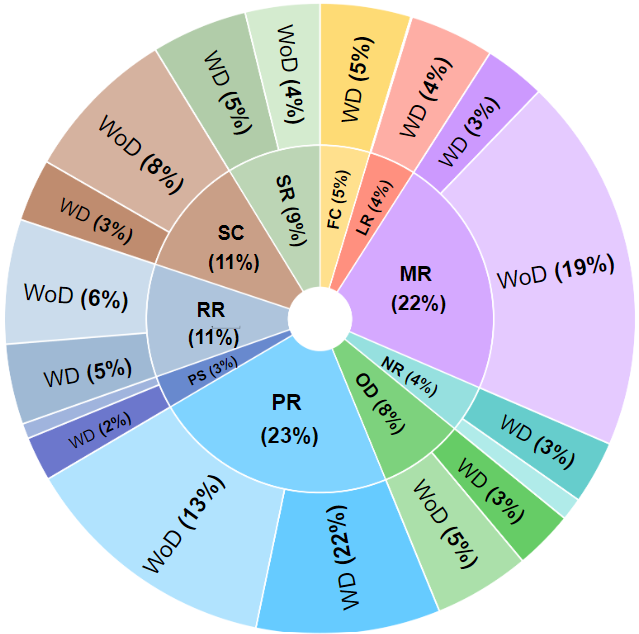}
        % \vspace{-2mm}
        \caption*{(a) Dataset categorization}
    \end{minipage}
    \begin{minipage}[t]{0.49\linewidth}
        \centering
        \includegraphics[
        width = 7 cm, height= 7.5 cm
        ]{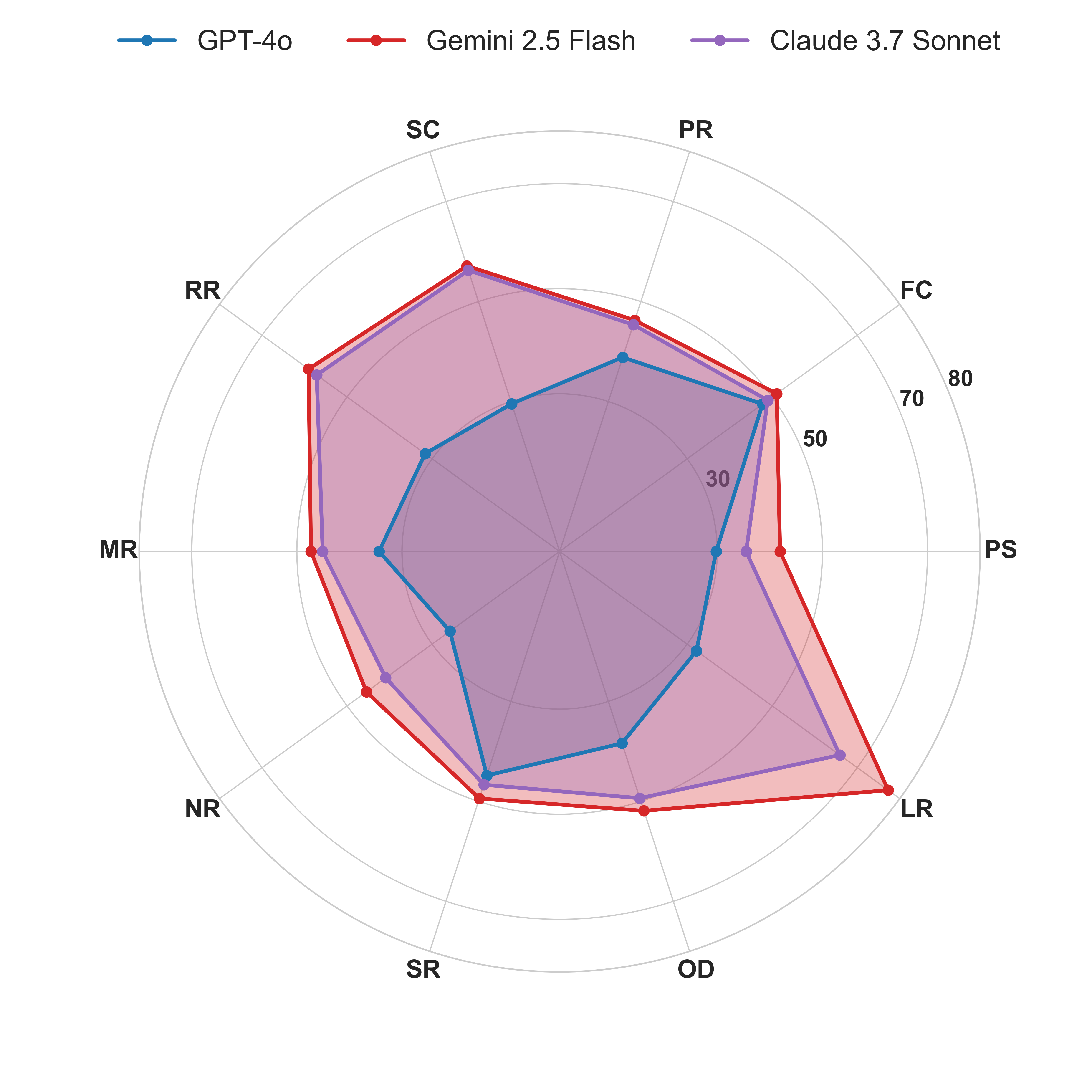}
        % \vspace{-6mm}
        \caption*{(b) Results on closed source models}
    \end{minipage}
    
    % \vspace{-3mm}
    \caption{An overview of \dataset's distribution and difficulty (a) exhibits the per-category split of the 5000 questions in the dataset, along with the split of \textit{with diagram} (WD) and \textit{without diagram} (WoD) for that category ; (b) Compares the per-category performance of various MLLMs.}
    % \vspace{-3mm}
    \label{fig:hor_2figs_1cap}
\end{figure}

\section{Related Work}
\label{sec:related_work}

The development of MLLMs builds on the progress of LLMs \citep{touvron2023llama,touvron2023llama2,OpenAI2023ChatGPT,albert24mixtral} and large vision models \citep{kirillov2023segment,zhang2023personalize,zhang2023prompt,zhang2023learning}. 
These models extend LLMs to handle a wider range of tasks across multiple modalities, including 2D images ~\citep{li2022blip,instructblip,alayrac2022flamingo,li2023mimic}, 3D point clouds ~\citep{guo2023point,xu2023pointllm}, audio ~\citep{han2023imagebind,su2023pandagpt}, and video ~\citep{zhang2023video,chen2023videollm}. Notable examples like OpenAI's GPT-4V ~\citep{openai2023gpt4v} and Google's Gemini ~\citep{team2023gemini} demonstrate advanced visual reasoning capabilities, setting new benchmarks in the multimodal space.

As MLLMs rapidly advance ~\citep{li2023multimodal}, there is a growing need for benchmarks that evaluate mathematical problem-solving in visual contexts. 
Existing benchmarks, such as GeoQA~\citep{Chen2021GeoQAAG}, VQA~\citep{goyal2017making}, and UniGeo~\citep{chen2022unigeo}, focus mostly on geometric problems. 
Other efforts target skills in abstract scenes, geometry diagrams, charts, and synthetic images~\citep{chen2022unigeo,masry2022chartqa}.
Recent datasets also assess external knowledge, commonsense reasoning, and scientific or medical understanding~\citep{zhang2023pmc}. 
MathVista~\citep{Lu2023MathVistaEM} expands multimodal math tasks, while MMMU~\citep{yue2023mmmu} focuses on college-level problems. Prior work evaluates LLMs across diverse domains like QA, mathematics, and science~\citep{bubeck2023sparks,nori2023capabilities}, while recent research~\citep{zhang2023lost} explores whether models like GPT-4V perform vision and language tasks independently or together.

\begin{table*}[t!]
\small
\centering
\begin{adjustbox}{width=\linewidth}
\begin{tabular}{lllllllllllc}
\toprule
\multicolumn{1}{l|}{\textbf{Category}} &
  \multicolumn{1}{c}{PS} &
  \multicolumn{1}{c}{FC} &
  \multicolumn{1}{c}{PR} &
  \multicolumn{1}{c}{SC} &
  \multicolumn{1}{c}{RR} &
  \multicolumn{1}{c}{MR} &
  \multicolumn{1}{c}{NR} &
  \multicolumn{1}{c}{SR} &
  \multicolumn{1}{c}{OD} &
  \multicolumn{1}{l|}{LR} &
  Overall \\ \midrule
\multicolumn{12}{c}{\textit{Full dataset}}                                                                                                    \\ \midrule
\multicolumn{1}{l|}{Questions with Diag.} & 114 & 233 & 472  & 160 & 206 & 157  & 162 & 246 & 151 & \multicolumn{1}{l|}{3}   & 1904                     \\
\multicolumn{1}{l|}{Questions w/o Diag.}  & 39  & 0   & 664  & 398 & 319 & 964  & 58  & 191 & 246 & \multicolumn{1}{l|}{217} & 3096                     \\
\multicolumn{1}{l|}{Total Questions}      & 153 & 233 & 1136 & 558 & 525 & 1121 & 220 & 437 & 397 & \multicolumn{1}{l|}{220} & 5000                     \\ \midrule
\multicolumn{12}{c}{\textit{testmini}}                                                                                                    \\ \midrule
\multicolumn{1}{l|}{Questions with Diag.} & 27  & 47  & 102  & 33  & 47  & 28   & 30  & 53  & 38  & \multicolumn{1}{l|}{0}   & 405                      \\
\multicolumn{1}{l|}{Questions w/o Diag.}  & 4   & 0   & 125  & 79  & 58  & 196  & 14  & 34  & 41  & \multicolumn{1}{l|}{44}  & 595                      \\
\multicolumn{1}{l|}{Total Questions}      & 31  & 47  & 227  & 112 & 105 & 224  & 44  & 87  & 79  & \multicolumn{1}{l|}{44}  & 1000                     \\ \midrule
\multicolumn{12}{c}{\textit{test-img}}                                                                                                    \\ \midrule
\multicolumn{1}{l|}{Total Questions}      & 60  & 122 & 248  & 84  & 108 & 82   & 85  & 129 & 79  & \multicolumn{1}{l|}{3}   & \multicolumn{1}{l}{1000} \\ \bottomrule
\end{tabular}
\end{adjustbox}
\caption{An overview of the per-category distribution of questions in the \textit{test}, \textit{testmini}, and \textit{test-img} splits of \dataset. \textit{testmini} and \textit{test-img} are 1000-instance subsets, aimed at faster and image-focused evaluations respectively. We also report the frequency of \textit{with diagram} and \textit{without diagram} questions for each category.}
\label{tab:all_dataset_stats}
\end{table*}

Existing extensive benchmarks ~\citep{fu2023mme,liu2023mmbench,Li2023SEEDBenchBM,xu2023lvlm} primarily focus on concrete, real-world problems within specific domains. 
These benchmarks often include comparatively simple diagram interpretation questions involving plots or mathematical questions related to geometry, which primarily evaluate models' abilities to parse information from a single image and solve problems using well-established logical principles and formulae. 
However, they do not sufficiently test models' capabilities in abstract visual reasoning, including spatial recognition, visual logic and puzzle solving, and pattern recognition.
This limitation represents a notable gap, as visual puzzle tasks require logical leaps that differ fundamentally from reasoning patterns over textual or linguistic problems. 
Moreover, spatial reasoning questions assess models' abilities to internalize and manipulate configurations in 3D space, as well as reason over spatial information and infer implicit relationships based on positional data. 
This category of questions aligns closely with human cognition and reasoning abilities, and evaluating model performance against human baselines on these questions reveals the substantial gap in reasoning abilities that models must bridge to approach human-comparable reasoning capability.
Our proposed dataset aims to address this gap by challenging and comprehensively evaluating previously underexplored model skills in categories where their performance still lags significantly behind human reasoning baselines. 
Additionally, we provide a detailed analysis of the strengths and weaknesses of these models across a wide range of categories and skills, shedding light on specific reasoning errors and their frequency of occurrence across categories and in comparison to one another.

\section{Curating \dataset}
\label{sec:dataset}

\dataset is curated mainly from questions directed at students taking the National Talent Search Examination, a nationwide competitive exam held by the National Council of Educational Research and Training  of India.
These questions and their solutions are created by experts in their fields and rigorously peer-reviewed, and thus contain minimal errors. 
These questions aim to assess Scholastic Aptitude (SAT), or the ability to recall domain-specific scientific and mathematical knowledge, as well as Mental Ability (MAT), or the ability to think logically and apply a range of analytical skills. 
We catalog the skills assessed by each sample along the categorization schema defined in Table \ref{tab:categ_definition}.

\subsection{Collection Pipeline}

To guarantee high-quality data, we manually collected image snippets and engineered a streamlined, automated framework for curation and annotation. 
Continuous human reviews were conducted throughout the process, ensuring quality and preventing error propagation.

\begin{itemize}
    \item \textbf{Step 1}: We generate a universally unique identifier (UUID) for a given question paper to identify all the questions curated from it. 
    \item \textbf{Step 2}: Annotators manually collected separate snippets of each question and their associated contextual information (including disconnected pieces) that apply to multiple questions.
    \item \textbf{Step 3}: An image merging script automatically identified and merged question images (in case the question gets split by pages) with their relevant context images.    
    \item \textbf{Step 4}: We used an LLM to transcribe the questions and their ground truth answers. 
    We also generate additional metadata, including category (\S \ref{sec:categs}), whether it contains a diagram 
    % (Fig \ref{fig:our_new_3_datasets})
    , and image description (\S \ref{sec:add_metadata_main}).
    A manual check was performed to ensure the quality of the generated metadata.

    \item \textbf{Step 5}: An annotation file, where each row corresponds to a question, is automatically created and populated.
 
\end{itemize}

\begin{table*}[!t]
\small
\centering
\begin{adjustbox}{width=\linewidth}
\begin{tabular}{llcc}
\toprule
\multicolumn{1}{c|}{\multirow{2}{*}{\textbf{Category name}}} &
  \multicolumn{1}{c|}{\multirow{2}{*}{\textbf{Definition}}} &
  \multirow{2}{*}{\textbf{\begin{tabular}[c]{@{}c@{}}Avg \\ len\end{tabular}}} &
  \multirow{2}{*}{\textbf{\begin{tabular}[c]{@{}c@{}}Max \\ len\end{tabular}}} \\
\multicolumn{1}{c|}{} &
  \multicolumn{1}{c|}{} &
   &
   \\ \midrule
\multicolumn{1}{l|}{Perspective Shift (PS)} &
  \multicolumn{1}{l|}{\begin{tabular}[c]{@{}l@{}}A figure is given and the solver is instructed to morph it \\ according to the instructions (flip, mirror image, rotate, etc.)\end{tabular}} &
  18.60 &
  59 \\ \midrule
\multicolumn{1}{l|}{Figure Completion (FC)} &
  \multicolumn{1}{l|}{\begin{tabular}[c]{@{}l@{}}A figure is given with an arrangement of numbers or characters \\ such that their relationship to one another based on their position \\ in the figure is consistent. The goal is to complete the figure and \\ identify the element missing from a marked position.\end{tabular}} &
  23.97 &
  364 \\ \midrule
\multicolumn{1}{l|}{Pattern Recognition (PR)} &
  \multicolumn{1}{l|}{\begin{tabular}[c]{@{}l@{}}This requires the understanding of a  one-to-one relationship \\ or pattern and replicating that pattern. For example, given the \\ relationship between a and b, determining the equivalent of \\ b to c. Questions involving substituting characters and \\ operations in a pre-defined pattern fall into this category.\end{tabular}} &
  31.98 &
  391.4 \\ \midrule
\multicolumn{1}{l|}{Sequence Completion (SC)} &
  \multicolumn{1}{l|}{\begin{tabular}[c]{@{}l@{}}Given a sequence of numbers or  figures, this question \\ involves finding the sequentially next element in a series.\end{tabular}} &
  30.22 &
  227 \\ \midrule
\multicolumn{1}{l|}{Relative Reasoning (RR)} &
  \multicolumn{1}{l|}{\begin{tabular}[c]{@{}l@{}}The question contains distinct data points and their relationship \\ with one another. The solver must extrapolate relationships that \\ may not be explicitly mentioned to answer the questions. \\ Questions involving Venn diagrams, family relations, or relative \\ positions given a reference point fall into this category.\end{tabular}} &
  27.22 &
  137 \\ \midrule
\multicolumn{1}{l|}{Mathematical Reasoning (MR)} &
  \multicolumn{1}{l|}{\begin{tabular}[c]{@{}l@{}}This question entails calculations of a mathematical nature, \\ such as solving a given equation.\end{tabular}} &
  25.61 &
  156 \\ \midrule
\multicolumn{1}{l|}{Numerical Reasoning (NR)} &
  \multicolumn{1}{l|}{\begin{tabular}[c]{@{}l@{}}Questions involving counting the number of elements \\ mentioned. The elements may be part of a single figure \\ or conform to a specified pattern.\end{tabular}} &
  15.63 &
  65 \\ \midrule
\multicolumn{1}{l|}{Spatial Reasoning} &
  \multicolumn{1}{l|}{\begin{tabular}[c]{@{}l@{}}These questions require the solver to visualize the context \\ and reason observationally to arrive at the answer.\end{tabular}} &
  27.67 &
  78 \\ \midrule
\multicolumn{1}{l|}{Odd One Out (OD)} &
  \multicolumn{1}{l|}{\begin{tabular}[c]{@{}l@{}}Given a set of elements, identify the element that is not like \\ the others.\end{tabular}} &
  26.64 &
  214 \\ \midrule
\multicolumn{1}{l|}{Logical Reasoning (LR)} &
  \multicolumn{1}{l|}{\begin{tabular}[c]{@{}l@{}}Questions involving simple logical reasoning such as \\ entailment and contradiction.\end{tabular}} &
  34.68 &
  144 \\ \midrule
\textbf{Overall} &
  \textbf{} &
  \textbf{27.68} &
  \textbf{391.4} \\ \bottomrule
\end{tabular}
\end{adjustbox}
\caption{An overview of our question categorization schema. Questions are categorized on the basis of the information provided in the question and the reasoning skills assessed.}
\label{tab:categ_definition}
\end{table*}

\subsection{Dataset categorization}\label{sec:categs}
We develop a categorization schema that catalogues questions on basis of the information provided and the type of reasoning assessed by the question.
Based on the continuous human evaluation during collection, we identify 10 distinct question categories.
We enumerate these categories along with their definitions in Table \ref{tab:categ_definition}.
We further distinguish between questions \textit{with diagram} and \textit{without diagram}. 
% The \textit{with diagram} questions are designed around the information presented in the diagrams (Fig \ref{fig:our_new_3_datasets}).
% We show examples of \textit{with diagram} and \textit{without diagram} questions for each category in \S \ref{examples_by_category}. 
The overall per-category distribution, along with the \textit{with diagram} and \textit{without diagram} split, is visualized in Figure \ref{fig:hor_2figs_1cap}.

\subsection{Additional metadata}
\label{sec:add_metadata_main}
The complexity of collected question images and the heavy presence of diagram-based reasoning tasks makes \dataset a challenging multi-modal benchmark. 
To make \dataset usable for both text and vision model evaluations, we provide transcriptions of questions and answers. 
To further facilitate text-based evaluation, we generate detailed, human-vetted text descriptions of attached diagrams such that a human could visualize the image based on this description 
% (Fig \ref{fig:our_new_3_datasets}). 
Results on text-only characterization of questions in our dataset can be found in \S \ref{sec:test-img}.

\subsection{Quality Assurance}

Following the collection and annotation process, we conduct a comprehensive quality check.
We discard samples that are [1] of low resolution, [2] outside the scope of the categories (Table \ref{tab:categ_definition}), or [3] missing vital information.
We also discard samples with noticeable watermarks and other visual noise that renders the sample illegible. 
Our subject-expert annotators rectify incorrectly-extracted ground truth answers. 
Concurrently, we verify that the questions belong to their assigned categories, and correct any observed misalignments therein.

\subsection{Division of the \textit{testmini} Subset.}
The final iteration of \dataset comprises 5000 questions. 
To enable faster model validation, we extract a 1000-instance subset, \textit{testmini}, using stratified sampling over all categories.
All quantitative results reported were obtained on this \textit{testmini} subset of \dataset.
We also create a \textit{test-img} question set, consisting solely of 1000 \textit{with diagram} questions, aimed at faster, focused assessment of models' visual comprehension. 
% Owing to the imbalance of \textit{with diagram} questions across categories, 
We use a random sampling strategy to create \textit{test-img} due to diagram imbalance. 
\footnote{All datasets (\textit{test}, \textit{testmini} and \textit{test-img}) will be publicly released}
For data distribution, see Table \ref{tab:all_dataset_stats}.
% Further details on data collection and annotation are available in \S \ref{sec:dataset_details_appendix}.

\section{Experiments}
\label{sec:experiments}

We conduct a systematic evaluation of existing MLLMs on \dataset. 
We first introduce the experimental setup in this section. 
Then we present our findings followed by multiple dataset analysis experiments. 
% Additional experimental results and qualitative examples are present in \S \ref{exp_setup} and \ref{qualitative}.
% Then, we detail the error analysis in \S \ref{sec:error_analysis} and qualitative results in \S \ref{qualitative}. 

\subsection{Experimental Setup}

\paragraph{Evaluation Models:}
We examine the performance of foundation models across two distinct categories on \dataset: (a) \textbf{Closed-source MLLMs}, represented by models like GPT-4o \texttt{(gpt-4o-2024-05-13)} 
% \citep{gpt4o},OpenAI O1 \texttt{(o1-preview-2024-09-12, o1-mini-2024-09-12)} \citep{openaio1}, 
Gemini-2.5 Flash \texttt{(Gemini-2.5-flash-002)} \citep{team2023gemini}, 
Claude-3.7 Sonnet \texttt{(claude-3-7-sonnet)} 
% \citep{Claude3S} and Claude 3 Haiku and Sonnet \texttt{(claude-3-sonnet-20240229, claude-3-haiku-20240307)} \citep{TheC3}
(b) \textbf{Open-source MLLMs}, such as 
LLaVA (v1.5-13B, v1.6-Mistral-7B, v1.6-Vicuna-13B) \citep{liu2023improvedllava}, 
LLaVA-v1.6-34B \citep{liu2024llavanext}, G-LLaVA (7B, 13B) \citep{gao2023g}, ShareGPT4V (7B, 13B) \citep{Chen2023ShareGPT4VIL} \& Qwen2-VL-2B-Instruct \citep{wang2024qwen2} (c) \textbf{Text Based LLMs}  Reka Flash \citep{ormazabal2024reka}, Llama-3 (70B) \citep{llama3modelcard}, Mistral Large \citep{AI_2024}.
We conduct experiments on open-source models using six NVIDIA A100 GPUs.
% Hyperparameters are available in \S \ref{exp_setup}.

\newcommand*{\minvalNorm}{28}%definetheminimumvalueonyourdataset
\newcommand*{\maxvalNorm}{60}%definethemaximumvalueinyourdataset!
\newcommand*{\opacity}{45}

\newcommand{\gradient}[1]{
\ifdimcomp{#1pt}{>}{\maxvalNormpt}{#1}{
\ifdimcomp{#1pt}{<}{\minvalNormpt}{#1}{
\pgfmathparse{int(round(100*(#1/(\maxvalNorm-\minvalNorm))-(\minvalNorm*(100/(\maxvalNorm-\minvalNorm)))))}
\xdef\tempa{\pgfmathresult}
\cellcolor{green!\tempa!red!\opacity}#1
}}
}

\newcommand{\gradientcell}[6]{
\ifdimcomp{#1pt}{>}{#3pt}{#1}{
\ifdimcomp{#1pt}{<}{#2pt}{#1}{
\pgfmathparse{int(round(100*(#1/(#3-#2))-(\minvalNorm*(100/(#3-#2)))))}
\xdef\tempa{\pgfmathresult}
\cellcolor{#5!\tempa!#4!#6}#1
}}
}

\newcommand{\gradientcellnorm}[1]{
\gradientcell{#1}{\minvalNorm}{\maxvalNorm}{red}{green}{\opacity}
}

\begin{table*}[!t]
\small
\centering
\begin{adjustbox}{width=\linewidth}
\begin{tabular}{lllllllllllr}
\toprule
\multicolumn{1}{l|}{\textbf{Category}} & \multicolumn{1}{c}{PS} & \multicolumn{1}{c}{FC} & \multicolumn{1}{c}{PR} & \multicolumn{1}{c}{SC} & \multicolumn{1}{c}{RR} & \multicolumn{1}{c}{MR} & \multicolumn{1}{c}{NR} & \multicolumn{1}{c}{SR} & \multicolumn{1}{c}{OD} & \multicolumn{1}{l|}{LR} & \multicolumn{1}{c}{\textbf{Overall}}\\\midrule
\multicolumn{12}{c}{\textit{Baseline}}\\\midrule
\multicolumn{1}{l|}{\textbf{Random chance}} & \multicolumn{1}{c}{9.68} & \multicolumn{1}{c}{4.26} & \multicolumn{1}{c}{6.61} & \multicolumn{1}{c}{9.82} & \multicolumn{1}{c}{9.52} & \multicolumn{1}{c}{9.82} & \multicolumn{1}{c}{15.91} & \multicolumn{1}{c}{6.90} & \multicolumn{1}{c}{7.59} & \multicolumn{1}{l|}{9.09} & 8.60\\
\multicolumn{1}{l|}{\textbf{Human eval}} & \multicolumn{1}{c}{51.08} & \multicolumn{1}{c}{70.57} & \multicolumn{1}{c}{61.82} & \multicolumn{1}{c}{69.35} & \multicolumn{1}{c}{69.84} & \multicolumn{1}{c}{76.64} & \multicolumn{1}{c}{58.71} & \multicolumn{1}{c}{62.64} & \multicolumn{1}{c}{64.98} & \multicolumn{1}{l|}{51.14} & 66.62\\\midrule
\multicolumn{12}{c}{\textit{Zero Shot Inference}}\\\midrule
\multicolumn{1}{l|}{\textbf{GPT-4o}} & \tcbox[colback=light_red]{29.79} & \tcbox[colback=light_red]{47.73} & \tcbox[colback=light_red]{38.84} & \tcbox[colback=light_red]{29.55} & \tcbox[colback=light_red]{31.65} & \tcbox[colback=light_red]{34.36} & \tcbox[colback=light_red]{25.81} & \tcbox[colback=light_red]{44.83} & \tcbox[colback=light_red]{38.39} & \multicolumn{1}{l|}{\tcbox[colback=light_red]{32.18}} &\gradientcellnorm{36.60}          
\\
\multicolumn{1}{l|}{\textbf{Gemini-2.5 Flash}} & \tcbox[colback=light_green]{41.94} & \tcbox[colback=light_green]{51.06} & \tcbox[colback=light_green]{46.26} & \tcbox[colback=light_green]{57.14} & \tcbox[colback=light_green]{59.05} & \tcbox[colback=light_green]{47.32} & \tcbox[colback=light_green]{45.45} & \tcbox[colback=light_green]{49.43} & \tcbox[colback=light_green]{51.90} & \multicolumn{1}{l|}{\tcbox[colback=light_green]{77.27}} & \gradientcellnorm{51.20}\\
\multicolumn{1}{l|}{\textbf{Claude-3.7 Sonnet}} & 35.48 & 48.94 & 45.37 & 56.25 & 57.14 & 45.09 & 40.91 & 46.67 & 49.37 & \multicolumn{1}{l|}{65.91} & \gradientcellnorm{48.60}\\\midrule
\multicolumn{12}{c}{\textit{Few Shot Inference}}\\\midrule
\multicolumn{1}{l|}{\textbf{GPT-4o}} & \tcbox[colback=light_red]{29.03} & \tcbox[colback=light_red]{14.89} & \tcbox[colback=light_red]{33.48} & \tcbox[colback=light_red]{38.39} & \tcbox[colback=light_red]{40.00} & \tcbox[colback=light_red]{40.18} & \tcbox[colback=light_red]{18.18} & \tcbox[colback=light_red]{36.78} & \tcbox[colback=light_red]{21.52} & \multicolumn{1}{l|}{\tcbox[colback=light_red]{50.00}} & \gradientcellnorm{34.60}
\\
\multicolumn{1}{l|}{\textbf{Gemini-2.5 Flash}} & \tcbox[colback=light_green]{48.39} & \tcbox[colback=light_green]{59.57} & \tcbox[colback=light_green]{47.58} & \tcbox[colback=light_green]{60.71} & \tcbox[colback=light_green]{61.90} & \tcbox[colback=light_green]{49.11} & \tcbox[colback=light_green]{52.27} & \tcbox[colback=light_green]{51.72} & \tcbox[colback=light_green]{54.43} & \multicolumn{1}{l|}{\tcbox[colback=light_green]{84.09}} & \gradientcellnorm{54.20}
\\
\multicolumn{1}{l|}{\textbf{Claude-3.7 Sonnet}} & 41.94 & 53.19 & 46.26 & 58.93 & 59.05 & 46.88 & 47.73 & 49.43 & 51.90 & \multicolumn{1}{l|}{75.00} & \gradientcellnorm{51.40}
\\\midrule
\multicolumn{12}{c}{\textit{Chain-of-Thought Prompting Inference}}\\\midrule
\multicolumn{1}{l|}{\textbf{GPT-4o}} & \tcbox[colback=light_red]{21.28} & \tcbox[colback=light_red]{54.55} & \tcbox[colback=light_red]{41.96} & \tcbox[colback=light_red]{25.00} & \tcbox[colback=light_red]{27.85} & \tcbox[colback=light_red]{29.96} & \tcbox[colback=light_red]{9.68} & \tcbox[colback=light_red]{40.95} & \tcbox[colback=light_red]{41.07} & \multicolumn{1}{l|}{\tcbox[colback=light_red]{33.33}} & \gradientcellnorm{35.00}\\
\multicolumn{1}{l|}{\textbf{Gemini-2.5 Flash}} & 51.61 & \tcbox[colback=light_green]{65.96} & \tcbox[colback=light_green]{48.02} & \tcbox[colback=light_green]{64.29} & \tcbox[colback=light_green]{64.76} & \tcbox[colback=light_green]{49.55} & \tcbox[colback=light_green]{59.09} & \tcbox[colback=light_green]{57.47} & \tcbox[colback=light_green]{58.23} & \multicolumn{1}{l|}{\tcbox[colback=light_green]{93.18}} & \gradientcellnorm{57.00}\\
\multicolumn{1}{l|}{\textbf{Claude-3.7 Sonnet}} & \tcbox[colback=light_green]{54.84} & 55.32 & 46.70 & 61.61 & 63.81 & 47.77 & 50.00 & 55.17 & 54.43 & \multicolumn{1}{l|}{77.27} & \gradientcellnorm{53.90}\\\midrule
\multicolumn{12}{c}{\textit{Step Back Prompting Inference}}\\\midrule
\multicolumn{1}{l|}{\textbf{GPT-4o}} & \tcbox[colback=light_red]{12.77} & \tcbox[colback=light_red]{45.45} & \tcbox[colback=light_red]{42.41} & \tcbox[colback=light_red]{27.27} & \tcbox[colback=light_red]{31.65} & \tcbox[colback=light_red]{34.80} & \tcbox[colback=light_red]{16.13} & \tcbox[colback=light_red]{41.90} & \tcbox[colback=light_red]{41.07} & \multicolumn{1}{l|}{\tcbox[colback=light_red]{37.93}} & \gradientcellnorm{36.50}\\
\multicolumn{1}{l|}{\textbf{Gemini-2.5 Flash}} & \tcbox[colback=light_green]{48.39} & \tcbox[colback=light_green]{59.57} & \tcbox[colback=light_green]{47.58} & \tcbox[colback=light_green]{62.50} & \tcbox[colback=light_green]{64.76} & \tcbox[colback=light_green]{48.21} & \tcbox[colback=light_green]{54.55} & \tcbox[colback=light_green]{55.17} & \tcbox[colback=light_green]{58.23} & \multicolumn{1}{l|}{\tcbox[colback=light_green]{88.64}} & \gradientcellnorm{55.40}\\
\multicolumn{1}{l|}{\textbf{Claude-3.7 Sonnet}} & \tcbox[colback=light_green]{48.39} & 48.94 & 45.37 & 60.71 & 63.81 & 47.77 & 45.45 & \tcbox[colback=light_green]{55.17} & 53.16 & \multicolumn{1}{l|}{75.00} & \gradientcellnorm{51.90}\\\bottomrule
\end{tabular}
\end{adjustbox}
\caption{Results of closed-source LLMs on the \textit{testmini} split of \dataset. We report model results using the following prompting strategies: zero-shot inference, few-shot inference, Chain-of-Thought, and Step Back prompting. For each prompting setting, the \tcbox{highest} and \tcbox[colback=light_red]{lowest} scores achieved by a model per category are highlighted. In addition to model accuracy, we report a Random chance baseline (i.e. the accuracy of a model that randomly selects an option without visibility into the question, and a Human eval baseline, where we report the average scores of six human evaluators.)}
\label{tab:main_results_closed_source}
\end{table*}
 %The min, mid and max values
\newcommand*{\minvalNormOSS}{4.9}% define the minimum value on your data set
\newcommand*{\maxvalNormOSS}{15.8}% define the maximum value in your data set!

% gradient function!
\newcommand{\gradientOSS}[1]{
    % The values are calculated linearly between \minvalNormOSS and \maxvalNormOSS
    \ifdimcomp{#1pt}{>}{\maxvalNormOSS pt}{#1}{
        \ifdimcomp{#1pt}{<}{\minvalNormOSS pt}{#1}{
            \pgfmathparse{int(round(100*(#1/(\maxvalNormOSS-\minvalNormOSS))-(\minvalNormOSS*(100/(\maxvalNormOSS-\minvalNormOSS)))))}
            \xdef\tempa{\pgfmathresult}
            \cellcolor{green!\tempa!red!\opacity} #1
    }}
}
%======================================
% gradient function single cell! 
\newcommand{\gradientcellOSS}[6]{
    % The values are calculated linearly between \minvalNormOSS and \maxvalNormOSS
    \ifdimcomp{#1pt}{>}{#3 pt}{#1}{
        \ifdimcomp{#1pt}{<}{#2 pt}{#1}{
            \pgfmathparse{int(round(100*(#1/(#3-#2))-(\minvalNormOSS*(100/(#3-#2)))))}
            \xdef\tempa{\pgfmathresult}
            \cellcolor{#5!\tempa!#4!#6} #1
    }}  
}

\newcommand{\gradientcellOSSnorm}[1]{
% \gradientcellOSS{cell_val}{min_val}{max_val}{colorlow}{colorhigh}{opacity}
    \gradientcellOSS{#1}{\minvalNormOSS}{\maxvalNormOSS}{red}{green}{\opacity}
}

\begin{table*}[!t]
\small
\centering
\begin{adjustbox}{width=\linewidth}
\begin{tabular}{l|rrrrrrrrrr|r}
\toprule
\textbf{Model}              & PS    & FC    & PR    & SC    & RR    & MR    & NR    & SR    & OD   & LR    & \textbf{Overall} \\ 
\midrule
\textbf{Qwen2 VL (2B) Instruct} & 9.38  & 2.13  & \tcbox[colback=light_red]{6.17} & 6.25 & 8.57 & \tcbox[colback=light_red]{3.57} & 4.55  & \tcbox[colback=light_red]{4.60} & 8.86 & \tcbox[colback=light_red]{2.27} & \gradientcellOSSnorm{5.60}            \\
\textbf{LLaVA-v1.6 Mistral (7B)} & 6.45  & 4.26  & 14.98 & \tcbox{14.29} & \tcbox{18.10} & \tcbox{15.18} & 9.09  & \tcbox{19.54} & \tcbox{22.78} & \tcbox{13.64} & \gradientcellOSSnorm{15.20}            \\
\textbf{G-LLaVA (7B)}            & 12.90 & \tcbox[colback=light_red]{0.00}  & 9.25  & \tcbox[colback=light_red]{3.57}  & \tcbox[colback=light_red]{5.71}  & 7.59  & 2.27  & \tcbox[colback=light_red]{4.60}  & \tcbox[colback=light_red]{3.80}  & 6.82  & \gradientcellOSSnorm{6.30}             \\ 
\textbf{ShareGPT4V (7B)}         & 6.45  & 10.64 & \tcbox{16.30} & 13.39 & 7.62  & 11.61 & 11.36 & 11.49 & 10.13 & 11.36 & \gradientcellOSSnorm{12.10}            \\
\textbf{LLaVA-v1.6 Vicuna (13B)} & 12.90 & 12.77 & 8.37  & 8.04  & 13.33 & 5.80  & \tcbox{15.91} & 6.90  & 13.92 & 4.55  & \gradientcellOSSnorm{9.10}             \\
\textbf{LLaVA 1.5 (13B)}         & \tcbox[colback=light_red]{3.23}  & 14.89 & 7.49  & 11.61 & 7.62  & 6.70  & 9.09  & 8.05  & 11.39 & \tcbox{13.64} & \gradientcellOSSnorm{8.70}             \\
\textbf{ShareGPT4V (13B)}        & 9.68  & 17.02 & 13.66 & 12.50 & 15.24 & 10.71 & 9.09  & 12.64 & 17.72 & 6.82  & \gradientcellOSSnorm{12.80}            \\
\textbf{G-LLaVA (13B)}            & \tcbox{13.67} & 2.33  & 11.12  & 5.69  & 7.98  & 10.23  & \tcbox[colback=light_red]{1.07}  & 6.70  & 5.76  & 7.98  & \gradientcellOSSnorm{8.26}             \\ 
\textbf{LLaVA-v1.6 (34B)} & 9.68  & \tcbox{25.33}  & 9.69 & 12.50 & 6.67 & 10.71 & 13.64  & 10.34 & 15.19 & 9.09 & \gradientcellOSSnorm{11.30}            \\
\bottomrule
\end{tabular}
\end{adjustbox}
\caption{Results of open-source MLLMs on the \textit{testmini} split of \dataset.
We report model results using zero shot inference. 
The \tcbox{highest} and \tcbox[colback=light_red]{lowest} scores achieved by a model in each category are highlighted.
}
\label{tab:main_results_open_source}
\end{table*}

\paragraph{Implementation Details} All reported results are on the $testmini$ subset. 
As a comparative baseline, we simulate random chance by selecting a random option for multiple-choice questions over 1000 trials. 
Additionally, the problems in \dataset were independently solved by the paper’s authors (four engineering graduates and two PhDs), serving as a human performance baseline.
We evaluate the benchmark using various prompting methods, including zero shot, few shot (2-shot), Chain-of-Thought \citep{wei2022chain}, and Step Back prompting \citep{zhengtake}. 
For multiple-choice questions, we use exact match for answer comparison.
The model inference prompts are structured to elicit a step-by-step solution, the final answer, and the corresponding option. 
% Details about these prompts are provided in \S \ref{prompts}.
As part of our analysis, we conducted three additional experiments: (1) analyzing model performance on the \textit{test-img} split, (2) converting the questions from \textit{test-img} into text, along with the transformation of diagrams into descriptions, and (3) evaluating OpenAI o1 models on questions without diagrams.

\subsection{Results}
\label{sec:results}

\paragraph{\textbf{Closed Source Models}}
Across various prompting strategies (Table \ref{tab:main_results_closed_source}), 
% Claude-3.7 Sonnet consistently emerged as the top performer, particularly excelling in complex tasks like SR,PR, and SC, with scores as high as 41.90\%. 
  Gemini-2.5 Flash performed best with these advanced prompts, achieving up to 57.00\% accuracy in Step Back Prompting, compared to 54.20\% in few shot.
Claude-3.7 Sonnet followed closely, especially in FC and PS questions, showing strong performance with zero shot and Step Back Prompting. 
GPT-4o Flash performed moderately across all categories but lacked dominance in any specific area.
% while Claude Haiku being the smallest of the closed sourced MLLMs,
% \kevin{Can better rephrase it?}
% consistently underperformed across all prompting strategies.
In terms of prompting strategies, Chain-of-Thought and Step Back Prompting enhanced the performance of top models like Claude-3.7 Sonnet and Gemini-2.5 Flash, allowing them to excel in tasks requiring structured reasoning and re-evaluation. 
Both strategies led to marked improvements over zero shot prompting, in categories like SR, PR, and LR.

\paragraph{\textbf{Open Source Models}}
Table \ref{tab:main_results_open_source} showcases the results of open-source MLLMs. 
LLaVA-v1.6-Mistral-7B model achieved the highest overall score of 15.2\%. 
% demonstrating remarkable performance across several categories. 
It excelled in OD (22.78\%), SR (19.54\%), RR (18.1\%), and MR (15.18\%)  indicating its proficiency in generating precise, coherent, and relevant responses, even for out-of-distribution samples.
The ShareGPT4V (13B) model exhibited the second-highest overall score of 12.8\%, with outstanding performance in the PR (13.66\%), SC (12.5\%), RR (15.24\%), MR (10.71\%), SR (12.64\%), and OD (17.72\%) categories. 
Other models, such as LlaVA-v1.6-Vicuna 13B, LlaVA-1.5 (13B), G-LLaVA (13B), and LlaVA-v1.6 (34B), exhibited varying levels of success across the different categories, highlighting their individual strengths and weaknesses in handling the diverse reasoning aspects tested by the dataset.

\paragraph{\textbf{Human Evaluation}}
To ascertain the difficulty of the dataset, we asked six graduate students specifically for the evaluation of human performance on \dataset.
We assigned questions from a specific problem category to each student. 
% This strategy prevented them from gleaning information from other questions within the category.  
They were asked to provide only the final answer without detailed reasoning, simulating zero-shot inference. 
% Therefore, we do not report the Chain-of-Thought evaluation results for human performance, alongside the Random Chance baseline.

 %The min, mid and max values
\newcommand*{\minvalNormIMG}{19}% define the minimum value on your data set
\newcommand*{\maxvalNormIMG}{35}% define the maximum value in your data set!

% gradient function!
\newcommand{\gradientIMG}[1]{
    % The values are calculated linearly between \minvalNormIMG and \maxvalNormIMG
    \ifdimcomp{#1pt}{>}{\maxvalNormIMG pt}{#1}{
        \ifdimcomp{#1pt}{<}{\minvalNormIMG pt}{#1}{
            \pgfmathparse{int(round(100*(#1/(\maxvalNormIMG-\minvalNormIMG))-(\minvalNormIMG*(100/(\maxvalNormIMG-\minvalNormIMG)))))}
            \xdef\tempa{\pgfmathresult}
            \cellcolor{green!\tempa!red!\opacity} #1
    }}
}
%======================================
% gradient function single cell! 
\newcommand{\gradientcellIMG}[6]{
    % The values are calculated linearly between \minvalNormIMG and \maxvalNormIMG
    \ifdimcomp{#1pt}{>}{#3 pt}{#1}{
        \ifdimcomp{#1pt}{<}{#2 pt}{#1}{
            \pgfmathparse{int(round(100*(#1/(#3-#2))-(\minvalNormIMG*(100/(#3-#2)))))}
            \xdef\tempa{\pgfmathresult}
            \cellcolor{#5!\tempa!#4!#6} #1
    }}  
}

\newcommand{\gradientcellIMGnorm}[1]{
% \gradientcellIMG{cell_val}{min_val}{max_val}{colorlow}{colorhigh}{opacity}
    \gradientcellIMG{#1}{\minvalNormIMG}{\maxvalNormIMG}{red}{green}{\opacity}
}

%%%%%%%%%%%%%%%%%%%

\newcommand*{\minvalNormIMGTO}{14.5}% define the minimum value on your data set
\newcommand*{\maxvalNormIMGTO}{16}% define the maximum value in your data set!

% gradient function!
\newcommand{\gradientIMGTO}[1]{
    % The values are calculated linearly between \minvalNormIMG and \maxvalNormIMG
    \ifdimcomp{#1pt}{>}{\maxvalNormIMGTO pt}{#1}{
        \ifdimcomp{#1pt}{<}{\minvalNormIMGTO pt}{#1}{
            \pgfmathparse{int(round(100*(#1/(\maxvalNormIMGTO-\minvalNormIMGTO))-(\minvalNormIMGTO*(100/(\maxvalNormIMGTO-\minvalNormIMGTO)))))}
            \xdef\tempa{\pgfmathresult}
            \cellcolor{green!\tempa!red!\opacity} #1
    }}
}
%======================================
% gradient function single cell! 
\newcommand{\gradientcellIMGTO}[6]{
    % The values are calculated linearly between \minvalNormIMG and \maxvalNormIMG
    \ifdimcomp{#1pt}{>}{#3 pt}{#1}{
        \ifdimcomp{#1pt}{<}{#2 pt}{#1}{
            \pgfmathparse{int(round(100*(#1/(#3-#2))-(\minvalNormIMGTO*(100/(#3-#2)))))}
            \xdef\tempa{\pgfmathresult}
            \cellcolor{#5!\tempa!#4!#6} #1
    }}  
}

\newcommand{\gradientcellIMGTOnorm}[1]{
% \gradientcellIMG{cell_val}{min_val}{max_val}{colorlow}{colorhigh}{opacity}
    \gradientcellIMGTO{#1}{\minvalNormIMGTO}{\maxvalNormIMGTO}{red}{green}{\opacity}
}

\begin{table*}[!t]
\small
\centering
\begin{adjustbox}{width=\linewidth}
\begin{tabular}{lrrrrrrrrrrr}
\toprule
\multicolumn{1}{l|}{\textbf{Category}}          & PS    & FC    & PR    & SC    & RR    & MR    & NR    & SR    & OD   & \multicolumn{1}{r|}{LR}     & \textbf{Overall} \\ \midrule
\multicolumn{12}{c}{\textit{MLLM Inference on Diagrams (Multi-modal)}}                                                                                            \\ \midrule

\multicolumn{1}{l|}{\textbf{GPT-4o}} & \tcbox[colback=light_red]{20.00} & \tcbox[colback=light_red]{20.49} & \tcbox[colback=light_red]{22.18} & \tcbox[colback=light_red]{19.05} & \tcbox[colback=light_red]{23.15} & \tcbox[colback=light_red]{20.73} & \tcbox[colback=light_red]{20.00} & \tcbox[colback=light_red]{17.05} & \tcbox[colback=light_red]{34.18} & \multicolumn{1}{r|}{\tcbox[colback=light_red]{66.67}} & \gradientcellIMGnorm{21.80} \\
\multicolumn{1}{l|}{\textbf{Gemini-2.5 Flash}} & 26.67 & \tcbox{34.43} & \tcbox{27.42} & \tcbox{36.90} & \tcbox{39.81} & \tcbox{35.37} & 22.35 & \tcbox{28.68} & \tcbox{41.77} & \multicolumn{1}{r|}{\tcbox{100.00}} & \gradientcellIMGnorm{32.10} \\
\multicolumn{1}{l|}{\textbf{Claude-3.7 Sonnet}} & \tcbox{43.34} & 33.61 & \tcbox{27.42} & 29.76 & 37.03 & 31.71 & \tcbox{34.12} & 26.36 & 37.98 & \multicolumn{1}{r|}{\tcbox{100.00}} & \gradientcellIMGnorm{26.20} \\ 
\midrule

\multicolumn{12}{c}{\textit{MLLM Inference on Diagram Descriptions (Text-only)}}                                                                                             \\ \midrule

\multicolumn{1}{l|}{\textbf{GPT-4o}} & \tcbox[colback=light_red]{26.67} & \tcbox[colback=light_red]{28.69} & \tcbox{29.44} & \tcbox[colback=light_red]{23.81} & 31.48 & \tcbox[colback=light_red]{34.15} & 30.59 & 29.46 & \tcbox[colback=light_red]{27.85} & \multicolumn{1}{r|}{\tcbox[colback=light_red]{33.33}} & \gradientcellIMGnorm{29.30} \\
\multicolumn{1}{l|}{\textbf{Gemini-2.5 Flash}} & 38.33 & \tcbox{34.43} & \tcbox[colback=light_red]{27.82} & \tcbox{35.71} & \tcbox[colback=light_red]{28.71} & 41.46 & \tcbox[colback=light_red]{25.88} & \tcbox[colback=light_red]{26.36} & 34.17 & \multicolumn{1}{r|}{\tcbox{100.00}} & \gradientcellIMGnorm{31.50} \\
\multicolumn{1}{l|}{\textbf{Claude-3.7 Sonnet}} & \tcbox{43.34} & 33.61 & 29.43 & 29.76 & \tcbox{40.74} & \tcbox{42.69} & \tcbox{42.36} & \tcbox{32.56} & \tcbox{45.57} & \multicolumn{1}{r|}{\tcbox{100.00}} & \gradientcellIMGnorm{33.50} \\ \midrule

\multicolumn{12}{c}{LLM Inference on Diagram Descriptions (Text-only)}                                                                          \\ \midrule
\multicolumn{1}{l|}{\textbf{Mistral Large}}     & \tcbox[colback=light_red]{15.00} & \tcbox[colback=light_red]{13.11} & \tcbox[colback=light_red]{11.29} & \tcbox[colback=light_red]{15.48} & \tcbox[colback=light_red]{18.52} & \tcbox[colback=light_red]{13.41} & \tcbox[colback=light_red]{9.41}  & \tcbox[colback=light_red]{17.83} & \tcbox[colback=light_red]{25.32} & \multicolumn{1}{r|}{33.33}  & \gradientcellIMGTOnorm{14.90}            \\
\multicolumn{1}{l|}{\textbf{Reka Flash}}        & \tcbox{16.67} & \tcbox{13.93} & \tcbox{12.10} & \tcbox{16.67} & \tcbox{19.44} & \tcbox{14.63} & \tcbox[colback=light_red]{9.41}  & \tcbox{18.60} & \tcbox{26.58} & \multicolumn{1}{r|}{33.33}  & \gradientcellIMGTOnorm{15.80}            \\
\multicolumn{1}{l|}{\textbf{Llama-3 (70B)}}       & \tcbox{16.67} & \tcbox{13.93} & 11.69 & \tcbox{16.67} & \tcbox{19.44} & \tcbox{14.63} & \tcbox{10.59} & \tcbox{18.60} & \tcbox{26.58} & \multicolumn{1}{r|}{33.33}  & \gradientcellIMGTOnorm{15.80}            \\ \bottomrule
\end{tabular}
\end{adjustbox}
% \caption{Results of visual comprehension ablation study \textit{test-img} split of \dataset. 
% For the given models, we conduct [1] multi-modal inference on questions containing diagrams, [2] unimodal inference with a detailed text description in place of the diagram. 
% For each setting, the \tcbox{highest} and \tcbox[colback=light_red]{lowest} scores achieved by a model per category are highlighted.
% Additionally, we report the performance of unimodal LLMs on these text-only questions.}
\caption{Visual comprehension ablation results on \textit{test-img}. We compare [1] multi-modal inference with diagrams and [2] unimodal inference using text descriptions. \tcbox{Highest} and \tcbox[colback=light_red]{lowest} scores per category are highlighted. Unimodal LLM performance on text-only questions is also reported.}
\label{tab:text_only_analysis}
\end{table*}

\subsection{Experimental Analysis}
\label{sec:analysis}

\paragraph{\textbf{MLLMs Rely More on Image Descriptions than Image}}
\label{sec:test-img} 
To evaluate the visual reasoning capabilities, we used \textit{test-img} subset, which contains questions with diagrams. 
Additionally, we generated a text-only version of \textit{test-img} by replacing all diagrams with detailed textual descriptions. 
Both experiments were carried out in a zero shot setting.
Our analysis reveals three key findings. First, we observed a noticeable decline in performance on \textit{test-img}, particularly for models like GPT-4o and Claude-3.7 Sonnet, compared to their results on the \textit{testmini} subset. 
This suggests that both models perform well on questions without diagrams, and their decreased accuracy on \textit{test-img} is largely due to the presence of diagram-based problems.
Second, when we replaced the diagrams in \textit{test-img} with text descriptions, the performance of all models improved by $\sim6\%$, indicating that the models struggle with diagrams and benefit from textual representations.
Finally, we evaluated popular text-only LLMs such as LLaMA-3 (70B), Reka Flash, and Mistral Large on the text-description version of \textit{test-img}. 
Their scores ($\sim 15\%$) were lower than those of the MLLMs ($\sim 27\%$), underscoring the advantage of multi-modal models in handling visually-grounded tasks.

\begin{figure*}[t!]
  \centering
  % \vspace{-2mm}
  \includegraphics[width=1.0\textwidth]{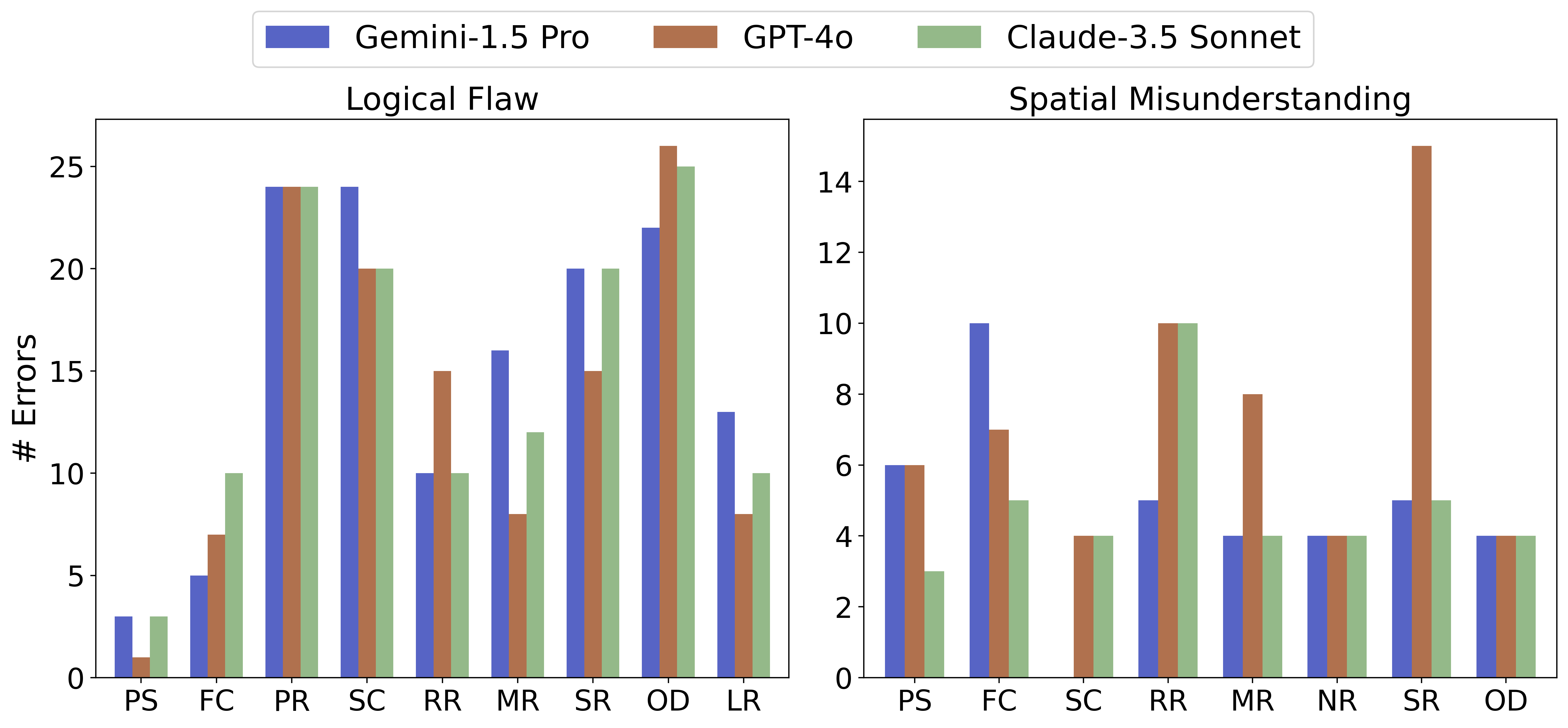}
  \vspace{-3mm}
  \caption{Frequency of LF and SM errors across different question categories. 
  We report per-model figures to enable a comparison of model abilities. 
  % LF and SM are the most common type of error we find. 
  They are most prevalent in the OD, PR, and SC categories of questions, owing to the amount of logical leaps and visual reasoning required by these questions.}
  % \vspace{-4mm}
\label{fig:category_wise_error_analysis}
\end{figure*}

\paragraph{\textbf{A Closer Look at Model Errors}} \label{sec:error_analysis}

We analysed 203 samples where all three state-of-the-art MLLMs (Claude-3.7 Sonnet, GPT-4o and Gemini-2.5 Flash) gave incorrect answers on \textit{testmini}.
Based on the manual inspection of the responses, we identified 7 types of errors that MLLMs make.
% (Table \ref{tab:error_analysis_table}). 
% The total error distribution of all three models is present in Table \ref{tab:error_analysis}. 
% Qualitative examples for category-wise errors are present in \S \ref{qualitative}.
The most common error on this dataset was Logical Flaw (LF), occurring in nearly $\sim 60\%$ of incorrect samples. 
Spatial Misunderstanding (SM), which involves a lack of understanding of diagram structure and content, was a close second ($\sim$ 20\%).
Figure \ref{fig:category_wise_error_analysis} shows the category-wise distribution of the two types of error.
These errors were most prevalent in OD, PR, and SC category of questions, as making uncommon logical leaps and fully comprehending visuals is integral to solving these. 
Furthermore, in questions involving extrapolation over multiple weakly connected data points, models came to conclusions that contradicted earlier data, indicating a lack of information retention.
Finally, we found that models fell into identical fallacious reasoning patterns, e.g. assuming that a pattern holds across each row when a pattern is replicated across columns. 
The category with the highest \% of shared errors was PR, where we observed that GPT, Gemini, and Claude followed the same incorrect reasoning structure on nearly 80\% of the analysed samples.
Thus, despite their differences, in practice we see that MLLMs share the same strengths and shortcomings.
% For more details, see \S \ref{sec:extended_error}.
% \ua{revise.}

% \paragraph{Evaluation of OpenAI o1 models}
% To understand the capabilities of recent text-only reasoning models (o1-preview and o1-mini), we evaluate these models on 595 text-only questions. 
% We also present human baseline scores on these questions. 
% These results are presented in Table \ref{tab:o1_case_study}. 
% o1-preview ($\sim 67\%$) scores competitively with human performance ($\sim 68\%$), while o1-mini ($\sim 58\%$) lags behind the human baseline by 10\%. 

\section{Conclusion}
In this work, we introduce \dataset, a benchmark designed to systematically analyze the mathematical reasoning capabilities of state-of-the-art models in visually complex scenarios. 
Our evaluation of 14 prominent foundation models highlights that significant advancements have been made, especially with the GPT-4o and Claude-3.7 Sonnet models. 
However, a substantial gap of $\sim10\%$ still exists between Gemini-2.5 Flash, the best-performing model, and human performance. 
This disparity sets a clear direction for future research, emphasizing the need for models that can seamlessly integrate mathematical reasoning with visual comprehension. 
Moreover, our analysis of model reasoning errors and experiments on samples containing diagrams and their textual representations offer valuable insights for future investigations.

\clearpage

\bibliographystyle{unsrt}
\bibliography{neurips_2025}

\appendix
\section*{Appendix}

\section*{Appendix Overview}
\begin{itemize}
    % \item Section~\ref{limit}: Limitation and Future Work
    \item Section~\ref{extended_related_work}: Extended Related Work
    \item Section~\ref{sec:dataset_creation_full}: Dataset creation
    \item Section~\ref{exp_setup}: Additional Experimental Details
    \item    Section~\ref{sec:extended_analysis}: Extended Analysis
    % Section~\ref{sec:extended_error}: More details on Error Analysis
    \item Section~\ref{sec:extended_error}: Qualitative Error Analysis
\end{itemize}

\begin{figure*}[t!]
  \centering
  % \vspace{-2mm}
  \includegraphics[width=1.0\textwidth]{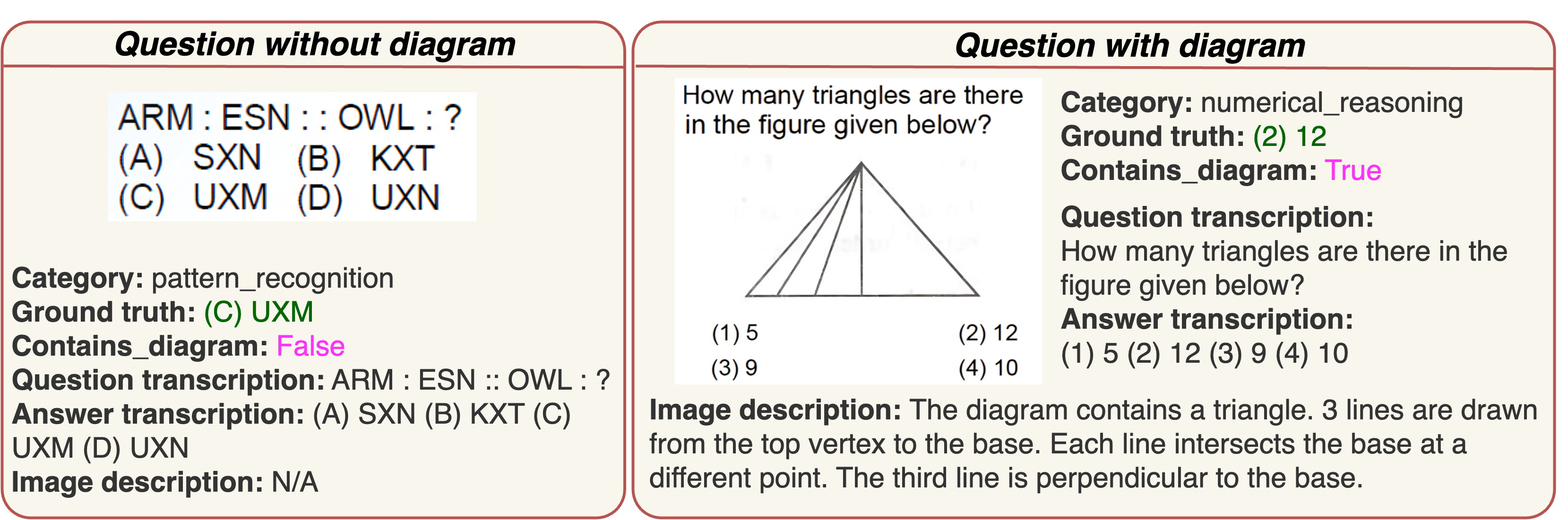}
  \vspace{-3mm}
  \caption{Examples of \textit{with diagram} and \textit{without diagram} questions. In addition to the question image, \dataset includes the metadata shown above. Question \textit{without diagram} is not present in \textit{test-img} while both kinds of questions will be present in \textit{testmini}.
  }
\label{fig:our_new_3_datasets}
\end{figure*}

\section{Extended Related Work} \label{extended_related_work}
% \textbf{[reviewer KHqw rebuttal]}
High-quality evaluation datasets and benchmarks are crucial for assessing the progress of machine learning models in solving real-world tasks~\citep{liao2021we}. Mathematical reasoning benchmarks have emerged as a significant focus area, posing challenges for large foundational models like Large Language Models (LLMs) and Multi-modal Large Language Models (MLLMs). Initial datasets addressed basic algebraic~\citep{hendrycksmath2021} and arithmetic~\citep{Roy2016SolvingGA} word problems with limited scope. Subsequent efforts, including MATH~\citep{hendrycksmath2021}, GSM8K~\citep{cobbe2021training}, MMLU~\citep{hendryckstest2021}, and others~\citep{zhou2023solving,yue2023mammoth,wang2024mathcoder,gao2023g,luo2023wizardmath}, expanded the range and quality of textual mathematical problems, establishing robust benchmarks for LLM evaluation.

Despite substantial mathematical reasoning encapsulated in visual modalities, most existing benchmarks~\citep{amini2019mathqa,cobbe2021training,mishra2022lila,frieder2023mathematical,lu2023dl4math} are textual only. Moreover, some datasets exhibit performance saturation, with GPT-4 achieving 92.0\% accuracy on GSM-8K~\citep{cobbe2021training}, a grade-school mathematics dataset. The rapid advancement of Large Multimodal Models (LMMs) necessitates robust multimodal benchmarks, as current benchmarks~\citep{antol2015vqa,kembhavi2016diagram,kahou2017figureqa,mathew2022infographicvqa} provide limited coverage of rigorous scientific domains crucial for general-purpose AI assistants.

While these benchmarks assess text-only mathematical reasoning, the rapid progress of MLLMs necessitates high-quality benchmarks for evaluating visual mathematical problem-solving. Prior attempts like GeoQA~\citep{Chen2021GeoQAAG}, while MathVista~\citep{Lu2023MathVistaEM} and MMMU~\citep{yue2023mmmu} incorporated various multimodal tasks and college-level questions, respectively.

MLLMs, building upon LLMs~\citep{touvron2023llama,touvron2023llama2,OpenAI2023ChatGPT,albert24mixtral,brown2020language} and large vision models~\citep{Radford2021LearningTV,kirillov2023segment,zhang2023personalize,zhang2023prompt,zhang2023learning}, have become increasingly prominent. They extend LLMs to diverse tasks and modalities, including 2D images~\citep{li2022blip,instructblip,alayrac2022flamingo,li2023mimic}, 3D point clouds~\citep{guo2023point,xu2023pointllm,hong20243d}, audio~\citep{han2023imagebind,su2023pandagpt}, and video~\citep{zhang2023video,chen2023videollm}. Noteworthy examples like OpenAI's GPT-4V~\citep{openai2023gpt4v} and Google's Gemini~\citep{team2023gemini} exhibit exceptional visual reasoning capabilities, setting new benchmarks in multi-modal performance.

However, their closed-source nature hinders broader application and development of MLLMs. Concurrently, open-source MLLMs like LLaMA-Adapter~\citep{zhang2024llamaadapter,gao2023llamaadapterv2}, LLaVA~\citep{liu2023llava,liu2024llavanext,liu2023improvedllava}, MiniGPT-4~\citep{zhu2023minigpt,chen2023minigpt}, mPLUG-Owl~\citep{ye2023mplugowl2}, Qwen-VL~\citep{bai2023qwen}, InternLM-XComposer~\citep{dong2024internlm}, and SPHINX~\citep{lin2023sphinx,gao2024sphinx} have been explored, leveraging CLIP~\citep{Radford2021LearningTV} for image encoding and LLaMA~\citep{touvron2023llama} for multi-modal instruction tuning, advancing MLLMs' visual understanding and generalization.

Despite comprehensive benchmarks~\citep{fu2023mme,liu2023mmbench,Li2023SEEDBenchBM,xu2023lvlm} for general visual instruction-following scenarios, the specific potential of MLLMs for visual mathematical problem-solving remains under-explored. Prior studies like VQA~\citep{antol2015vqa,goyal2017making}, VizWiz~\citep{gurari2018vizwiz}, and ParsVQA-Caps~\citep{mobasher101parsvqa} evaluate LMMs' general visual question answering abilities on open-ended image queries. Additionally, works have assessed LMMs' specific skills beyond natural scenes, such as abstract shapes~\citep{antol2015vqa,lu2021iconqa,ji2022abstract}, geometry diagrams~\citep{seo2015solving,lu2021inter,chen2022unigeo,cao2022augmented}, charts~\citep{methani2020plotqa,masry2022chartqa,kahou2017figureqa,chang2022mapqa,kafle2018dvqa}, documents~\citep{singh2019towards,mathew2022infographicvqa,liu2023hidden}, synthetic images~\citep{dahlgren2022clevr,li2023super,bitton2023breaking}, external knowledge~\citep{schwenk2022okvqa,shah2019kvqa}, commonsense reasoning~\citep{zellers2019recognition,yin2021broaden}, scientific knowledge~\citep{lu2022learn,kembhavi2017you,kembhavi2016diagram}, and medical understanding~\citep{zhang2023pmc,lau2018dataset}.

Generative foundation models like GPT-3~\citep{brown2020language}, GPT-4~\citep{gpt4}, Claude~\citep{claude2}, LLaMA~\citep{touvron2023llama}, and LLaMA-Adapter~\citep{llamaadapter2023} can solve various downstream tasks~\citep{wei2022emergent} without task-specific fine-tuning. Prior work has evaluated their text-based abilities in QA, math, medicine, coding, and science~\citep{bubeck2023sparks,nori2023capabilities,chen2021evaluating,fu2023codeapex,sun2023scieval,wang2023scibench,huang2023c,huang2022language,liu2023agentbench,llamaadapter2023}. Some work focused on specialized pretraining for improved visual math and chart reasoning, like PixStruct~\citep{lee2023pix2struct}, MatCha~\citep{liu2022matcha}, and UniChart~\citep{masry2023unichart}. On the vision-language front, models like LLaVA~\citep{liu2023llava}, miniGPT4~\citep{zhu2023minigpt}, InstructBLIP~\citep{instructblip}, Flamingo~\citep{alayrac2022flamingo,awadalla2023openflamingo}, LLaMA-Adapter V2~\citep{gao2023llamaadapterv2}, and Multimodal Bard~\citep{google2023bard} leverage paired~\citep{schuhmann2022laion,sharma2018conceptual,lin2014microsoft} and interleaved~\citep{zhu2023multimodal} image-text data. Additionally, specialized versions like LLaVAR~\citep{zhang2023llavar,ye2023mplug} emphasize document understanding and math comprehension. Recent works like Visit-Bench~\citep{bitton2023visit}, LVLM-eHub~\citep{yu2023mm}, MMBench~\citep{liu2023mmbench,xu2023lvlm,shao2023tiny} assess these models' instruction-following and reasoning capabilities.

Large language models (LLMs) have demonstrated remarkable reasoning abilities, further enhanced by approaches like chain-of-thought (CoT)~\citep{wei2022chain}, program-of-thought (PoT)~\citep{chen2022program}, and inductive reasoning~\citep{wang2023hypothesis, tan2023large}. The feasibility of using LLMs to solve the Abstraction and Reasoning Corpus (ARC) challenge has been verified using zero-shot, few-shot, and context-grounded prompting~\citep{tan2023large}.

OpenAI's GPT-4V, the multimodal version of GPT-4, exhibits promising performance in vision-language reasoning. However, a fine-grained study of its strengths and limitations is still lacking. Recent work~\citep{zhang2023lost} explores whether large multimodal models (LMMs) like GPT-4V execute vision and language tasks consistently or independently, contributing pioneering efforts in this field.

\section{Dataset creation}
\label{sec:dataset_creation_full}
\subsection{ Collection Pipeline}
\label{sec:dataset_details_appendix}
To ensure high-quality samples, all data samples were manually collected as image snippets from publicly available websites. 
% including:
% \begin{itemize}
%     \item https://scert.kerala.gov.in/
% \item https://scertharyana.gov.in/
% \item https://rajeduboard.rajasthan.gov.in/
% \item https://scholarships.wbsed.gov.in/
% \end{itemize}
We developed a flexible, highly automated data curation framework to streamline the process and standardize collection and annotation. 
Continuous human reviews were conducted between steps in the pipeline to maintain quality and prevent error propagation.

\begin{itemize}
    \item Step 1: A universally unique identifier (UUID) was generated for each question paper to track all curated questions. This step also updated a shared record containing details of the paper and the annotator’s alias, enabling efficient assignment of questions for peer review. 
    \item Step 2: Annotators manually collected individual snippets of each question, along with contextual information relevant to multiple questions. For questions requiring additional context, snippets were labeled accordingly, and only legible, relevant questions (focused on Mental Ability or Scholastic Ability in mathematics) were included to maintain dataset integrity.
    \item Step 3: An image-merging script automatically identified and merged split question images or context snippets (based on the naming convention) using open-source image processing tools\footnote{https://opencv.org/}. This resulted in a single image for each sample in the \dataset set of questions used to test models.
    \item Step 4: The next module in the pipeline created and automatically populated an annotation file, where each row corresponded to a collected sample. Columns included the paper\_id (UUID from Step 1), question number, and image path.
    \item Step 5: Using an answer key or solution set, LLM-powered transcription extracted the ground truth answers for each question. Extracted answers were mapped to the corresponding annotation rows, followed by a manual check to ensure alignment with the provided solution and correctness.
    \item LLM-based categorization, followed by human verification, was performed to obtain question categories. 
    Table \ref{tab:prompt_categorization} is the prompt used for the categorization of questions into various problem types. 
    Figures \ref{fig:dataset_categories_1}, \ref{fig:dataset_categories_2}, \ref{fig:dataset_categories_3}, \ref{fig:dataset_categories_4}, \ref{fig:dataset_categories_5}, \ref{fig:dataset_categories_6}, \ref{fig:dataset_categories_7}, \ref{fig:dataset_categories_8}, \ref{fig:dataset_categories_9},
\ref{fig:dataset_categories_10} demonstrate examples from each question category defined in Table \ref{tab:categ_definition}.
\end{itemize}

\section{Additional experiment details} \label{exp_setup}

\subsection{Hyperparameters} 
% The following hyperparameters were used as part of our experiments. 
% Gemini-1.5 Pro - $temperature:1, top_p: 0.95, top_k: 64, max\_output\_tokens: 8192, response\_mime\_type: text/plain$.

% GPT-4o - $top_p: 0.1, temperature: 1, max\_output\_tokens: 4096, stream: False$.

% Claude Family - $top_p: 0.1, temperature: 1, max\_output\_tokens: 4096, stream: False$.

% Open Source Models - $max\_new\_tokens: 3600, temperature: 0.7, top_p: 0.3, num\_beams: 1$.

\begin{table*}[ht]
\centering
\begin{tabular}{>{\bfseries}l|l}
\toprule
\textbf{Model} & \textbf{Hyperparameters} \\ \midrule
Gemini-1.5 Pro & 
\begin{tabular}[c]{@{}l@{}}
temperature: 1, top\_p: 0.95, top\_k: 64,\\
max\_output\_tokens: 8192,\\
response\_mime\_type: text/plain
\end{tabular} \\ \midrule
GPT-4o & 
\begin{tabular}[c]{@{}l@{}}
top\_p: 0.1, temperature: 1,\\
max\_output\_tokens: 4096, stream: False
\end{tabular} \\ \midrule
Claude Family & 
\begin{tabular}[c]{@{}l@{}}
top\_p: 0.1, temperature: 1,\\
max\_output\_tokens: 4096, stream: False
\end{tabular} \\ \midrule
Open Source Models & 
\begin{tabular}[c]{@{}l@{}}
max\_new\_tokens: 3600, temperature: 0.7,\\
top\_p: 0.3, num\_beams: 1
\end{tabular} \\ \bottomrule
\end{tabular}
\caption{Hyperparameters used in the experiments}
\label{table:hyperparameters}
\end{table*}
The experimental hyperparameters are enumerated in Table \ref{table:hyperparameters}.
Furthermore, Table \ref{tab:model_details} provides the source repositories and model cards for the various models used in our experiments.
Table \ref{tab:open_source_other_prompts_results} shows the performance of open-source models across categories using two additional prompting strategies: $Chain\text{-}of\text{-}Thought$ and $Step\text{-}back$.
Table \ref{tab:error_analysis} shows the total count of error analysis sample distribution that was conducted.
  
\begin{table*}[!t]
\small
\centering
\begin{adjustbox}{width=\linewidth}
\begin{tabular}{l@{\hspace{0.5cm}}c@{\hspace{1cm}}p{0.4\textwidth}}
    \toprule
    \textbf{Model} & \textbf{\makecell{Release\\ Time}} & \textbf{\makecell[c]{Source}} \\
    \midrule
    
    GPT-4o~\cite{gpt4o}     &   2023-03    & \url{https://platform.openai.com/} \\
    \midrule
    
    Claude 3 family~\cite{Claude3S,TheC3}     &   2023-03    & \url{https://www.anthropic.com/news/claude-3-family} \\
    \midrule
    
    Gemini-1.5 Pro~\cite{team2023gemini}    & 2023-12    & \url{https://ai.google.dev/} \\
    \midrule
    
    \multirow{2}{*}{LLaVA-1.5~\cite{liu2023improvedllava}}         &    \multirow{2}{*}{2023-10}       & \url{https://huggingface.co/liuhaotian/llava-v1.5-13b}  \\
    \midrule
    
    \multirow{2}{*}{G-LLaVA~\cite{gao2023g}}       &   \multirow{2}{*}{2023-12}    &        \url{https://github.com/pipilurj/G-LLaVA/tree/main} \\
    \midrule
    
    \multirow{2}{*}{ShareGPT4V~\cite{Chen2023ShareGPT4VIL}}       &   \multirow{2}{*}{2023-11}    &        \url{https://github.com/ShareGPT4Omni/ShareGPT4V/blob/master/docs/ModelZoo.md#sharegpt4v-models} \\
    \midrule
    
    \multirow{1}{*}{LLaVA-NeXT~\cite{liu2024llavanext}}         &    \multirow{1}{*}{2024-01}       & \url{https://github.com/LLaVA-VL/LLaVA-NeXT} \\
    \midrule
    
    \multirow{2}{*}{Qwen2-VL~\cite{wang2024qwen2}}  &  \multirow{2}{*}{2024-01} &\url{https://huggingface.co/Qwen/Qwen2-VL-2B-Instruct} \\
    \bottomrule
    
    \end{tabular}
\end{adjustbox}
\caption{Models used to evaluated \dataset, along with their release dates and source repositories. We use both open-source and closed-source models for a comprehensive evaluation.}
\label{tab:model_details}
\end{table*}

\begin{table*}[!t]
\small
\centering
\begin{adjustbox}{width=\linewidth}
\begin{tabular}{lrrrrrrrrrrr}
\toprule
\multicolumn{1}{l|}{\textbf{Category}}              & PS    & FC    & PR    & SC    & RR    & MR    & NR    & SR    & OOO   & \multicolumn{1}{r|}{LR}    & \textbf{Overall} \\ \midrule
\multicolumn{12}{c}{\textit{Chain of Thought Inference}}                                                                                                                    \\ \midrule
\multicolumn{1}{l|}{\textbf{Qwen2 VL 2B Instruct}}  & 12.90 & 2.13  & 6.61  & 0.89  & 9.52  & 3.57  & 6.82  & 5.75  & 10.13 & \multicolumn{1}{r|}{4.55}  & 5.70             \\
\multicolumn{1}{l|}{\textbf{Llava v1.6 Mistral 7B}} & 12.90 & 8.51  & 15.86 & 15.18 & 20.00 & 15.63 & 11.36 & 21.84 & 25.32 & \multicolumn{1}{r|}{15.91} & 16.80            \\
\multicolumn{1}{l|}{\textbf{G-LLaVA 7B}}            & 16.13 & 0.00  & 9.69  & 4.46  & 5.71  & 8.04  & 4.55  & 5.75  & 3.80  & \multicolumn{1}{r|}{9.09}  & 7.00             \\
\multicolumn{1}{l|}{\textbf{ShareGPT4V 7B}}         & 9.68  & 19.15 & 16.74 & 14.29 & 8.57  & 12.05 & 13.64 & 12.64 & 8.86  & \multicolumn{1}{r|}{13.64} & 13.20            \\
\multicolumn{1}{l|}{\textbf{Llava v1.6 Vicuna 13B}} & 16.13 & 17.02 & 9.25  & 9.82  & 14.29 & 6.25  & 18.18 & 9.20  & 15.19 & \multicolumn{1}{r|}{9.09}  & 10.60            \\
\multicolumn{1}{l|}{\textbf{Llava v1.5 13B}}        & 6.45  & 17.02 & 8.37  & 12.50 & 8.57  & 7.14  & 11.36 & 9.20  & 12.66 & \multicolumn{1}{r|}{15.91} & 9.80             \\
\multicolumn{1}{l|}{\textbf{ShareGPT4V 13B}}        & 12.90 & 19.15 & 14.10 & 13.39 & 16.19 & 11.61 & 11.36 & 14.94 & 18.99 & \multicolumn{1}{r|}{11.36} & 14.10            \\
\multicolumn{1}{l|}{\textbf{G-LLaVA 13B}}           & 16.13 & 2.13  & 11.45 & 6.25  & 8.57  & 10.27 & 2.27  & 6.90  & 6.33  & \multicolumn{1}{r|}{9.09}  & 8.70             \\
\multicolumn{1}{l|}{\textbf{Llava v1.6 34B}}        & 12.90 & 25.53 & 10.13 & 0.89  & 7.62  & 10.71 & 15.91 & 10.34 & 16.46 & \multicolumn{1}{r|}{9.09}  & 10.5             \\ \midrule
\multicolumn{12}{c}{\textit{Step Back Inference}}                                                                                                                           \\ \midrule
\multicolumn{1}{l|}{\textbf{Qwen2 VL 2B Instruct}}  & 16.13 & 4.26  & 7.05  & 1.79  & 10.48 & 4.02  & 9.09  & 6.90  & 11.39 & \multicolumn{1}{r|}{6.82}  & 6.70             \\
\multicolumn{1}{l|}{\textbf{Llava v1.6 Mistral 7b}} & 16.13 & 6.38  & 16.74 & 14.29 & 20.95 & 14.29 & 13.64 & 21.84 & 26.58 & \multicolumn{1}{r|}{18.18} & 17.00            \\
\multicolumn{1}{l|}{\textbf{G-LLaVA 7B}}            & 12.90 & 0.00  & 9.25  & 3.57  & 5.71  & 7.59  & 2.27  & 4.60  & 3.80  & \multicolumn{1}{r|}{6.82}  & 7.30             \\
\multicolumn{1}{l|}{\textbf{ShareGPT4V 7B}}         & 16.13 & 23.40 & 16.30 & 15.18 & 10.48 & 11.61 & 15.91 & 10.34 & 6.33  & \multicolumn{1}{r|}{15.91} & 13.50            \\
\multicolumn{1}{l|}{\textbf{Llava v1.6 Vicuna 13B}} & 19.35 & 14.89 & 10.13 & 8.04  & 13.33 & 6.70  & 20.45 & 10.34 & 16.46 & \multicolumn{1}{r|}{11.36} & 11.00            \\
\multicolumn{1}{l|}{\textbf{Llava 1.5 13B}}         & 12.90 & 14.89 & 8.37  & 13.39 & 7.62  & 7.59  & 13.64 & 8.05  & 13.92 & \multicolumn{1}{r|}{20.45} & 10.30            \\
\multicolumn{1}{l|}{\textbf{ShareGPT4V 13B}}        & 9.68  & 17.02 & 13.66 & 15.18 & 18.10 & 12.05 & 13.64 & 12.64 & 17.72 & \multicolumn{1}{r|}{15.91} & 14.30            \\
\multicolumn{1}{l|}{\textbf{G-LLaVA 13B}}           & 19.35 & 4.26  & 11.89 & 7.14  & 9.52  & 10.71 & 4.55  & 8.05  & 7.59  & \multicolumn{1}{r|}{11.36} & 9.70             \\
\multicolumn{1}{l|}{\textbf{Llava v1.6 34B}}        & 16.13 & 27.66 & 10.57 & 1.79  & 8.57  & 11.16 & 18.18 & 11.49 & 17.72 & \multicolumn{1}{r|}{11.36} & 11.50            \\ \bottomrule
\end{tabular}
\end{adjustbox}
\caption{Results of open-source MLLMs on the \textit{testmini} split of \dataset. 
We report model results using Chain-of-Thought, and Step Back prompting methods. 
}
\label{tab:open_source_other_prompts_results}
\end{table*}

\begin{table*}[!t]
\small
\centering
\begin{adjustbox}{width=\linewidth}
\begin{tabular}{l|l|rrr}
\toprule
\textbf{Error Name} &
  \textbf{Definition} &
  \multicolumn{1}{l}{\textbf{Gemini}} &
  \multicolumn{1}{l}{\textbf{GPT}} &
  \multicolumn{1}{l}{\textbf{Claude}} \\ \midrule
Incomplete (IC) &
  Model generated incomplete solution, or output hit token limit &
  5.91 &
  5.42 &
  0.49 \\
Logical Flaw (LF) &
  \begin{tabular}[c]{@{}l@{}}Reasoning step violated established logical rules or real-world \\ principles (such as equality or cardinality)\end{tabular} &
  60.09 &
  55.66 &
  56.16 \\
Memory Flaw (MF) &
  \begin{tabular}[c]{@{}l@{}}Model forgets information provided in the question or \\ earlier in the solution\end{tabular} &
  8.37 &
  10.34 &
  11.82 \\
\begin{tabular}[c]{@{}l@{}}Spatial \\ Misunderstanding (SM)\end{tabular} &
  \begin{tabular}[c]{@{}l@{}}Model misunderstands spatial relations or “misreads” specific \\ details of given image.\end{tabular} &
  18.72 &
  19.70 &
  17.73 \\
\begin{tabular}[c]{@{}l@{}}Calculation \\ Error (CE)\end{tabular} &
  \begin{tabular}[c]{@{}l@{}}Model commits a mathematical error, or substitutes the \\ wrong value in an equation.\end{tabular} &
  2.46 &
  1.48 &
  5.91 \\
Misalignment (MG) &
  \begin{tabular}[c]{@{}l@{}}Model reasons correctly, but concludes the answer incorrectly \\ (eg. identifying the pattern but selecting the wrong option )\end{tabular} &
  5.42 &
  8.87 &
  8.37 \\ \bottomrule
\end{tabular}
\end{adjustbox}
\caption{ The types of errors found in model reasoning patterns. The errors are defined to be mutually distinct and leave very little room for ambiguity. We also report the frequency of these errors for each model (Gemini-2.5 Flash, Claude-3.7 Sonnet, GPT-4o) over the 203 questions analysed.}
    \label{tab:error_analysis_table}
\end{table*}
\begin{table*}[!t]
\small
\centering
\begin{adjustbox}{width=\linewidth}
\begin{tabular}{llllllllllll}
\toprule
\multicolumn{1}{l|}{\textbf{Error Type}} &
  \multicolumn{1}{r}{\textbf{PS}} &
  \multicolumn{1}{r}{\textbf{FC}} &
  \multicolumn{1}{r}{\textbf{PR}} &
  \multicolumn{1}{r}{\textbf{SC}} &
  \multicolumn{1}{r}{\textbf{RR}} &
  \multicolumn{1}{r}{\textbf{MR}} &
  \multicolumn{1}{r}{\textbf{NR}} &
  \multicolumn{1}{r}{\textbf{SR}} &
  \multicolumn{1}{r}{\textbf{OD}} &
  \multicolumn{1}{r|}{\textbf{LR}} &
  \multicolumn{1}{r}{\textbf{Overall}} \\ \midrule
\multicolumn{12}{c}{\textit{Gemini-2.5 Flash}}                                                                              \\ \midrule
\multicolumn{1}{l|}{Calculation Error (CE)}              & 1  & 0  & 0  & 0  & 0  & 3  & 1  & 0  & 0  & \multicolumn{1}{l|}{0}  & 5   \\
\multicolumn{1}{l|}{Incomplete (IC)}              & 1  & 0  & 0  & 1  & 7  & 2  & 1  & 0  & 0  & \multicolumn{1}{l|}{0}  & 12  \\
\multicolumn{1}{l|}{Logical Flaw (LF)}            & 3  & 3  & 20 & 23 & 10 & 10 & 0  & 20 & 20 & \multicolumn{1}{l|}{13} & 122 \\
\multicolumn{1}{l|}{Memory Flaw (MF)}                  & 0  & 1  & 3  & 0  & 3 & 1  & 4  & 5  & 0  & \multicolumn{1}{l|}{0}  & 17  \\
\multicolumn{1}{l|}{Misalignment (MG)}            & 3  & 0  & 0  & 4  & 0  & 0  & 0  & 0  & 4  & \multicolumn{1}{l|}{0}  & 11  \\
\multicolumn{1}{l|}{Spatial Misunderstanding (SM)}                 & 6  & 10 & 0  & 0  & 5  & 4  & 4  & 5  & 4  & \multicolumn{1}{l|}{0}  & 38  \\ \midrule
\multicolumn{1}{l|}{\textbf{Overall Errors}} & 14 & 14 & 23 & 28 & 25 & 20 & 10 & 28 & 28 & \multicolumn{1}{l|}{13} & 203 \\ \midrule
\multicolumn{12}{c}{\textit{GPT-4o}}                                                                                      \\ \midrule
\multicolumn{1}{l|}{Calculation Error (CE)}              & 1  & 0  & 0  & 0  & 0  & 1  & 1  & 0  & 0  & \multicolumn{1}{l|}{0}  & 3   \\
\multicolumn{1}{l|}{Incomplete (IC)}              & 0  & 3  & 0  & 4  & 0  & 3  & 1  & 0  & 0  & \multicolumn{1}{l|}{0}  & 11  \\
\multicolumn{1}{l|}{Logical Flaw (LF)}            & 1  & 7  & 20 & 20 & 11 & 8  & 0  & 15 & 23 & \multicolumn{1}{l|}{6}  & 113 \\
\multicolumn{1}{l|}{Memory Flaw (MF)}                  & 0  & 0  & 6  & 0  & 5  & 6  & 4  & 0  & 0  & \multicolumn{1}{l|}{0}  & 21  \\
\multicolumn{1}{l|}{Misalignment (MG)}            & 6  & 0  & 0  & 4  & 0  & 1  & 0  & 0  & 0  & \multicolumn{1}{l|}{7}  & 18  \\
\multicolumn{1}{l|}{Spatial Misunderstanding (SM)}                 & 4  & 4  & 0  & 1  & 5 & 8  & 4  & 10 & 4  & \multicolumn{1}{l|}{0}  & 40  \\ \midrule
\multicolumn{1}{l|}{\textbf{Overall Errors}} & 12 & 14 & 26 & 29 & 20 & 27 & 10 & 25 & 27 & \multicolumn{1}{l|}{13} & 203 \\ \midrule
\multicolumn{12}{c}{\textit{Claude-3.7 Sonnet}}                                                                           \\ \midrule
\multicolumn{1}{l|}{Calculation Error (CE)}              & 1  & 0  & 0  & 0  & 0  & 9 & 1  & 0  & 1  & \multicolumn{1}{l|}{0}  & 12  \\
\multicolumn{1}{l|}{Incomplete (IC)}              & 0  & 0  & 0  & 1 & 0  & 0  & 0  & 0  & 0  & \multicolumn{1}{l|}{0}  & 1   \\
\multicolumn{1}{l|}{Logical Flaw (LF)}            & 3  & 8 & 18 & 25 & 10 & 6 & 1  & 10 & 22 & \multicolumn{1}{l|}{10} & 114 \\
\multicolumn{1}{l|}{Memory Flaw (MF)}                  & 1  & 0  & 2  & 0  & 10 & 1  & 4  & 5  & 0  & \multicolumn{1}{l|}{1}  & 24  \\
\multicolumn{1}{l|}{Misalignment (MG)}            & 6  & 0  & 0  & 8  & 0  & 0  & 0  & 0  & 0  & \multicolumn{1}{l|}{3}  & 17  \\
\multicolumn{1}{l|}{Spatial Misunderstanding (SM)}                 & 3  & 5  & 2  & 4  & 10 & 4  & 2  & 2  & 4  & \multicolumn{1}{l|}{0}  & 36  \\ \midrule
\multicolumn{1}{l|}{\textbf{Overall Errors}} & 14 & 13 & 22 & 38 & 30 & 20 & 8 & 17 & 27 & \multicolumn{1}{l|}{14} & 203 \\ \bottomrule
\end{tabular}
\end{adjustbox}
\caption{Type of errors made by Gemini-2.5 Flash, GPT4-o, and Claude-3.7 Sonnet over various question categories.}
\label{tab:error_analysis}
\end{table*}

%%% proportion of SM 
% \begin{table}
% \begin{tabular}{lll}
% \toprule
% Category          & Diagrams & SM \\ \midrule
% Figure completion & 100      & 28(sonnet)-55(gemini)\%                  \\
% Perspective shift & 66       & 25(sonnet)-50(gemini,gpt4o)\%                  \\
% Spatial reasoning & 55       & 16(gemini,sonnet)-50(gpt4o)\%                  \\ \bottomrule
% \end{tabular}
% \end{table}

\begin{table*}[!t]
\small
\centering
\begin{adjustbox}{width=\linewidth}
\begin{tabular}{l|rrrrrrrrrr|r}
\toprule
\textbf{Category} & PS    & FC    & PR    & SC    & RR    & MR    & NR    & SR    & OOO   & LR    & \textbf{Overall} \\ \midrule
\textbf{Human 1}  & 45.16 & 80.85 & 52.86 & 69.64 & 74.29 & 67.86 & 52.27 & 60.92 & 72.15 & 40.91 & 63.10            \\
\textbf{Human 2}  & 41.94 & 53.19 & 45.81 & 80.36 & 84.76 & 85.71 & 75.00 & 77.01 & 75.95 & 40.91 & 69.10            \\
\textbf{Human 3}  & 67.74 & 63.83 & 86.78 & 54.46 & 61.90 & 80.80 & 72.73 & 44.83 & 79.75 & 40.91 & 70.70            \\
\textbf{Human 4}  & 64.52 & 78.72 & 85.90 & 47.32 & 43.81 & 80.80 & 47.73 & 68.97 & 56.96 & 56.82 & 68.30            \\
\textbf{Human 5}  & 45.16 & 87.23 & 45.81 & 79.46 & 80.00 & 75.00 & 54.55 & 60.92 & 51.90 & 75.00 & 65.10            \\
\textbf{Human 6}  & 41.94 & 59.57 & 53.74 & 84.82 & 74.29 & 69.64 & 50.00 & 63.22 & 53.16 & 52.27 & 63.40            \\ \bottomrule
\end{tabular}
\end{adjustbox}
\caption{Per-category accuracy scores achieved by six human evaluators. The average human accuracy over all categories is 66.62\%.}
\label{tab:human_results}
\end{table*}

\subsection{Prompts for inference}\label{prompts}

The various prompts are detailed in this section.  Table \ref{tab:image_description} is the prompt used for generating the alternate image description of the question which is present as detailed in the additional metadata section \S \ref{sec:add_metadata_main}. 
Table \ref{tab:zero_shot_prompt}, \ref{tab:cot_prompt}, \ref{tab:step_back_prompt} show cases the zero shot prompt, Chain of thought and Step back prompt for inference on \dataset respectively.
Table \ref{tab:answer_options_list_extraction} shows the answer extraction prompt from the MLLM response
Table \ref{tab:text_based_inference} shows the text based inference for Analysis \ref{tab:text_only_analysis}.

\begin{table*}[!ht]
\centering
\begin{boxedminipage}{\columnwidth}
You are given a question designed to test a student on mathematical or logical reasoning. These questions can be categorized based on the skills and techniques used to solve them. \\These are the categories of questions.\\ \\ 

Mathematical reasoning: this question purely requires calculations of a mathematical nature. This includes solving a straightforward equation.\\ 

Pattern recognition: this requires the understanding of a one-to-one relationship or pattern and replicating that pattern. For example, given the relationship between a and b, determining the equivalent of b to c. Questions involving substituting characters and operations in a pre-defined pattern fall into this category.\\ 

Sequence completion: given a sequence of numbers or figures, this question involves finding the sequentially next element in a series.\\ 

Figure completion: You are given a figure with an arrangement of numbers or characters such that their relationship to one another based on their position in the figure is consistent. Th goal is to complete the figure and identify the element missing from a marked position. \\ 

Odd one out: given a set of elements, identify the element that is not like the others. \\ 

Spatial reasoning: questions involving reasoning observationally and visualizing the question in order to arrive at the answer.\\ 

Perspective shift: Questions where a figure is given and you are instructed to morph it according to the instructions (flip, mirror image, rotate, etc)\\ 

Numerical reasoning: questions involving counting the number of elements mentioned. The elements may be part of a single figure or conform to a specified pattern, but solving these questions requires counting.\\ 

Relative reasoning: the question contains distinct data points, and solving the questions requires understanding the relationships between all data points and extrapolating relationships that are not explicitly mentioned. Questions involving venn diagrams, family relations, or relative positions given a reference point fall into this category.\\ 

Logical reasoning: Questions involving simple logical reasoning such as entailment and contradiction.\\ \\ Now, observe the following question.\\ \\ Using the categorization schema explained above, classify this question into a category. Provide a detailed explanation. Output a JSON with the key "question" containing a transcript of the question, "category" containing the classification category, and "explanation" containing the reasoning for assigning the question to this category, and "contains diagram" which should be True or False depending on whether there is a diagram provided in the question.
\end{boxedminipage}
\caption{Prompt used for categorization of question of image.}
\label{tab:prompt_categorization}
\end{table*}

\begin{table*}[h!]
\centering
\begin{boxedminipage}{\columnwidth}
You are given a mathematical question involving a diagram. You are an accessibility reader for the blind. Output a detailed text description describing the diagram.\\

Example description: "description": "The diagram contains a circle, triangle, and rectangle overlapping. The circle is the topmost figure, the triangle is figure with the lowest base. The rectangle top cuts through the circle and triangle, while its lower side only passes through the triangle. The portion of the circle that does not overlap with any other figure contains the number 10. The intersection between circle and triangle contains the number 12. The intersection of only the circle and rectangle contains the number 5. The area where all 3 figures intersect contains 20. The area of the rectangle that interacts with no other figure contains 14. The area of the intersection between only the rectangle and triangle contains 17. Finally, the area of the triangle does not intersect with any other figures contains the number 16. Outside these figures are text labels and arrows. The arrow labeled Teacher points to the circle. The arrow labeled Doctor points to the rectangle. The arrow labeled Musician points to the triangle."\\ 

Now, generate a similarly comprehensive text description for the diagram in this question. \\ 

Image:{{image}}\\

Remember, the description must be detailed enough that the user can recreate the diagram exactly as shown based on the description alone. Do not add any information or make assumptions that are not explicitly mentioned in the image.\\ 

Output a JSON with the key "description" whose value is the generated description. Output only the JSON. Go!
\end{boxedminipage}
\caption{Prompt used to generated detailed textual description of diagrams.}
\label{tab:image_description}
\end{table*}

\begin{table*}[h!]
\centering
\begin{boxedminipage}{\columnwidth}
Common Prefix: "You are given a question to solve below:\\
This question requires skills and reasoning related to {{category}}. Definition: {{category definition}}.\\ 
This question has a list of options : {{answer range}}.\\ 
Your output must be a valid JSON." \\
        
Zeroshot Prompt: "Q1: Provide a step by step solution to this question.\\ Q2: What is the answer to this question? Remember, the answer must be present in the given list of answer options\\ Q3: Which is the option from {{answer range}} that corresponds to the answer above? Output only the option and nothing else.\\ Output a JSON with the keys Q1, Q2, Q3 with their answers." \\

Common postfix: "Remember, your output must be a valid JSON in this format:{'Q1':<answer>,'Q2':<answer>,'Q3':<answer>} If your JSON is incomplete, incorrectly delimited or badly formatted, you will be destroyed. Output the valid JSON and nothing else. Go!"
\end{boxedminipage}
\caption{Prompt for zero shot inference}
\label{tab:zero_shot_prompt}
\end{table*}

\begin{table*}[h!]
\centering
\begin{boxedminipage}{\columnwidth}
Common Prefix: "You are given a question to solve below:\\
This question requires skills and reasoning related to {{category}}. Definition: {{category definition}}.\\ 
This question has a list of options : {{answer range}}.\\ 
Your output must be a valid JSON." \\
        
CoT Prompt: Now answer the following questions.\\Q1: What is the list of variables and their values provided in the questions?\\Q2: What is the variable that needs to be solved for?\\Q3: What information that is not present in the question, can you infer from the given variables?\\Q4: Provide a step-by-step solution with reasoning to obtain the answer to this question. Provide the solution at each step.\\Q5: What is the answer to this question? Remember, the answer must be present in the given list of answer options.\\Q6: Which is the option from {{answer range}} that corresponds to the answer above? Output only the option and nothing else.\\\\ Output a JSON with the keys Q1, Q2, Q3, Q4, Q5, Q6 with their answers. \\

Common postfix: "Remember, your output must be a valid JSON in this format:{'Q1':<answer>,'Q2':<answer>,'Q3':<answer>} If your JSON is incomplete, incorrectly delimited or badly formatted, you will be destroyed. Output the valid JSON and nothing else. Go!"
\end{boxedminipage}
\caption{Prompt for Chain-of-Thought inference}
\label{tab:cot_prompt}
\end{table*}

\begin{table*}[h!]
\centering
\begin{boxedminipage}{\columnwidth}
Common Prefix: "You are given a question to solve below: This question requires skills and reasoning related to {{category}}. Definition: {{category definition}}.\\ 
This question has a list of options : {{answer range}}. Your output must be a valid JSON." \\       
Step back category prompt: \\
\textbf{Mathematical Reasoning:} "Q1: What is the relation of all given variables to one another? How is each variable related to the missing value?\\Q2: Which are the mathematical operations involved in solving a question like this?" \\
\textbf{Pattern Recognition:} "Q1: What is the pattern being followed in this question? Provide an example.\\Q2: Which are the elements in this question that follow this pattern?" \\
\textbf{Sequence Completion:} "Q1: What is a numerical sequence?\\Q2: What is the relationship between previous and subsequent elements in a sequence? What is the relationship between elements in the sequence present in this question?" \\
                
Figure Completion: "Q1: How do you approach a figure completion problem? \\Q2: What is the information you have and the missing information? What are their spatial relationships to one another?" \\

Odd one out: "Q1: How do you identify an odd element out of a set?\\Q2: Describe the elements in this set. Now ,what do almost all of these elements have in common?" \\
                
Spatial Reasoning: "Q1: What are the spatial manipulations that occur in this question? Eg. unfolding, folding, 2D to 3D reconstruction, etc.\\Q2: Given the original question image, how can you visualize the resulting image after the manipulations mentioned in the question? Explain in detail." \\

Perspective Shift: "Q1: What are the attributes of an image that is flipped, rotated, or its mirror image? What differentiates the result of these operations from the original image?\\Q2: Which of these operations apply in this image, and in what order?"\\

Numerical Reasoning: "Q1: What is the information you are given? What do you need to find out? How can you arrive at this number? Q2: What are the main points of concern in solving such a question? How can you ensure that you do not under or over estimate the final number?"\\

Relative Reasoning: "Q1: What is the information you are given? What are the relationships of the given data points to one another? What is the information you need to discover? Which data points are directly or indirectly related to the missing variable? Explain in detail. \\Q2: What principles of relational logic do you need to apply to this question?"\\

Logical Reasoning: "Q1: what are the principle of logical reasoning involved in solving this question? Q2: What is the information provided in this question? What is the objective of this question?" \\

Meta Prompt: Step back category prompt. Q3: Based on the above information, provide a step-by-step solution to the question in the image. Q4: What is the answer to this question? Remember, the answer must be present in the given list of answer options  Q5: Which is the option from {{answer range}} that corresponds to the answer above? Output only the option and nothing else.\\ Output a JSON with the keys Q1, Q2, Q3, Q4, Q5 with their answers.

\end{boxedminipage}
\caption{Per-category and meta-prompts for Step Back prompt inference}
\label{tab:step_back_prompt}
\end{table*}

\begin{table*}[ht!]
\centering
\begin{boxedminipage}{\columnwidth}
You are given a mathematical question with a list of multiple choice answers. You are an accessibility reader for the blind. Transcribe the textual part of the question, and the list of answer options provided.\\Example: {'question':'How many triangles are present in this diagram?','answer list':'(A) 23 (B) 21 (C) 29 (D) 34'}\\Now, generate a question and answer list transcript for the question in the image.\\Output a JSON with the keys "question" and "answer list" as described. Output only the JSON. Go!
\end{boxedminipage}
\caption{Prompt to transcribe list of answer options from question image}
\label{tab:answer_options_list_extraction}
\end{table*}

\begin{table*}[ht!]
\centering
\begin{boxedminipage}{\columnwidth}
You are given a question to solve below:\\ \\ This question requires skills and reasoning related to {{category}}. This question contains a diagram that is crucial to solving the question whose textual description as been provided. Definition: {{category definition}}. Problem: {{extracted question}}. Diagram: {{image description}} {{extracted answer list}} \\ Q1: Provide a step by step solution to this question.\\ Q2: What is the answer to this question? Remember, the answer must be present in the given list of answer options\\ Q3: Which is the option from {{answer range}} that corresponds to the answer above? Output only the option and nothing else.\\ Output a JSON with the keys Q1, Q2, Q3 with their answers.\\Remember, your output must be a valid JSON in this format:{'Q1':<answer>,'Q2':<answer>,'Q3':<answer>} If your JSON is incomplete, incorrectly delimited or badly formatted, you will be destroyed. Output the valid JSON and nothing else. Go!
\end{boxedminipage}
\caption{Prompt for text-only inference.}
\label{tab:text_based_inference}
\end{table*}

\begin{table*}[ht!]
\small
\centering
\begin{adjustbox}{width=\linewidth}
\begin{tabular}{l|llllllllll|l}
\toprule
\textbf{Model} & \textbf{FC} & \textbf{LR} & \textbf{MR} & \textbf{NR} & \textbf{OD} & \textbf{PR} & \textbf{PS} & \textbf{RR} & \textbf{SC} & \textbf{SR} & \textbf{Overall} \\ \midrule
\textbf{Gemini-1.5 Pro}    & 24.03       & 37.27       & 34.61       & 30.00       & 36.27       & 27.11       & 16.34       & 30.29       & 29.39       & 32.49       & 30.68            \\
\textbf{GPT-4o}            & 22.32       & 55.91       & 34.61       & 40.91       & 47.86       & 27.82       & 19.61       & 32.00       & 25.81       & 47.37       & 34.16            \\
\textbf{Claude-3 sonnet}   & 21.46       & 43.64       & 25.51       & 32.73       & 33.25       & 25.09       & 22.88       & 33.14       & 27.60       & 27.23       & 28.06            \\
\textbf{Claude Haiku}      & 19.31       & 24.09       & 20.87       & 28.64       & 30.23       & 23.50       & 23.53       & 20.57       & 25.09       & 22.43       & 23.28            \\
\textbf{Claude-3.5 Sonnet} & 29.18       & 79.09       & 35.59       & 50.91       & 53.65       & 27.82       & 45.75       & 32.76       & 31.18       & 51.03       & 38.42            \\ \bottomrule
\end{tabular}
\end{adjustbox}
\caption{Results on the entire \dataset dataset}
\label{tab:full_dataset_performance}
\end{table*}

\begin{table*}[ht!]
\small
\centering
\begin{adjustbox}{width=\linewidth}
\begin{tabular}{l|llllllllll|l}
\toprule
\textbf{Model}             & \textbf{FC} & \textbf{LR} & \textbf{MR} & \textbf{NR} & \textbf{OD} & \textbf{PR} & \textbf{PS} & \textbf{RR} & \textbf{SC} & \textbf{SR} & \textbf{Overall} \\ \midrule
\textbf{Gemini 1.5 Pro}    & 27.45       & 33.99       & 35.95       & 28.76       & 32.68       & 23.53       & 13.07       & 33.33       & 29.39       & 33.99       & 29.21            \\
\textbf{GPT 4o}            & 19.61       & 58.82       & 32.68       & 43.14       & 45.75       & 30.07       & 22.88       & 28.76       & 24.18       & 48.37       & 35.42            \\
\textbf{Claude 3 sonnet}   & 19.61       & 45.75       & 22.22       & 35.95       & 35.95       & 22.22       & 19.61       & 36.60        & 26.14       & 28.76       & 29.28            \\
\textbf{Claude Haiku}      & 22.88       & 20.92       & 19.61       & 30.07       & 28.76       & 24.84       & 22.22       & 21.57       & 26.14       & 20.92       & 23.79            \\
\textbf{Claude 3.5 sonnet} & 26.14       & 82.35       & 38.56       & 47.71       & 55.56       & 26.80        & 49.02       & 29.41       & 30.07       & 52.29       & 43.79            \\ \bottomrule
\end{tabular}
\end{adjustbox}
\caption{Results on a 153-sample set of each category, showing model scores on a balanced distribution across question categories.}
\label{tab:balanced_categ_results}
\end{table*}

\section{Extended Analysis}
\label{sec:extended_analysis}
\subsection{Additional inference results}
In this section, we show inference results from additional experiments to further illustrate model performance on \dataset.
The results of open-source models on \textit{test-mini} is shown in Table \ref{tab:open_source_other_prompts_results}.
We also document model performance on the full \dataset dataset (5000 questions) in Table \ref{tab:full_dataset_performance}.
Additionally, we create a 153-sample set for each category to form a category-balanced subset of \dataset, on which we show model performance in Table \ref{tab:balanced_categ_results}.

\subsection{Reliance on diagram descriptions}
\label{sec:diagramdescr}

\begin{table*}[ht!]
\small
\centering
\begin{adjustbox}{width=\linewidth}
\begin{tabular}{l|llllllllll|l}
\toprule
{} &     \textbf{FC} &     \textbf{LR} &     \textbf{MR} &     \textbf{NR} &     \textbf{OD} &     \textbf{PR} &     \textbf{PS} &     \textbf{RR} &     \textbf{SC} &     \textbf{SR} &  \textbf{Overall} \\
\midrule
\textbf{Claude Haiku}   &  23.40 &  25.00 &  24.55 &  18.18 &  35.44 &  29.52 &  48.39 &  28.57 &  40.18 &  34.48 &       30.00 \\
\textbf{GPT 4o}  &  42.55 &  52.27 &  40.18 &  45.45 &  35.44 &  46.26 &  70.97 &  33.33 &  53.57 &  60.92 &       45.60 \\
\textbf{Gemini 1.5 Pro} &  44.68 &  54.55 &  41.96 &  59.09 &  56.96 &  28.63 &  29.03 &  42.86 &  38.39 &  37.93 &       40.50 \\
\textbf{Claude-3 Sonnet}   &  42.55 &  47.73 &  36.16 &  50.00 &  62.03 &  29.52 &  29.03 &  42.86 &  38.39 &  37.93 &       39.00 \\
\textbf{Claude-3.5 Sonnet} &  46.81 &  56.82 &  42.86 &  59.09 &  37.97 &  47.14 &  83.87 &  36.19 &  58.04 &  63.22 &       49.00 \\
\bottomrule
\end{tabular}
\end{adjustbox}
\caption{Results of experiment setting combining diagram description, along with diagram image and question.}
\label{tab:additional_diagramdesc}
\end{table*}

In order to quantify the maximum performance gain achievable by providing diagram description to MLLMs, we conduct an additional experiment where we provide diagram description, question, and diagram for all questions in \textit{test-img}. The results are shown in Table \ref{tab:additional_diagramdesc}.

\section{Error Analysis} 
\label{sec:extended_error}
\subsection{Methodology} 
We leveraged 2 authors of this work to act as error evaluators independently and in parallel. 
Each evaluator has a graduate degree in Computer Science and experience in similar puzzle-solving.
Owing to the clear and mutually-exclusive definitions of error types, there is little ambiguity in identifying the error type of the incorrect responses.
Our measure of inter-evaluator agreement is Cohen's Kappa (K), found to be 0.9 - indicating near-unanimous agreement.
For questions where there was disagreement in evaluations, a consensus was reached after discussion.
\subsection{Quantitative Analysis}
We define the 6 types of errors found in model reasoning patterns and their frequency of occurrence in Table \ref{tab:error_analysis_table}.
Table \ref{tab:error_analysis} provides a detailed quantitative analysis of error type frequency per question category.  
Additionally, we analyse error patterns for the most-performant and least-performant open source models in Tables \ref{tab:best_perform} and \ref{tab:worst_perform} respectively.

\subsection{Qualitative Analysis}
\label{qualitative}

This section presents examples of the qualitative error analysis that was carried out. Figures \ref{fig:dataset_categories_1}, \ref{fig:dataset_categories_2},
\ref{fig:dataset_categories_3},
\ref{fig:dataset_categories_4},
\ref{fig:dataset_categories_5},
\ref{fig:dataset_categories_6},
\ref{fig:dataset_categories_7},
\ref{fig:dataset_categories_8},
\ref{fig:dataset_categories_9} and 
\ref{fig:dataset_categories_10}
contains examples of failures by three proprietary models viz. Gemini-1.5 Pro, GPT-4o, and Claude-3.5 Sonnet across all categories.

\begin{table}[ht!]
\centering
\begin{tabular}{l|r}
\toprule
\textbf{Row Labels} & \textbf{Percent} \\
\midrule
calc error    & 3.81   \\
incomplete    & 0.00   \\
logical flaw  & 63.55  \\
memory        & 2.11   \\
misalign      & 0.84   \\
spatial       & 29.66  \\
\bottomrule
\end{tabular}
\caption{Qwen2 VL (2B) Instruct (5) - Least performant open source model}
\label{tab:worst_perform}
\end{table}

\begin{table}[ht!]
\centering
\begin{tabular}{l|r}
\toprule
\textbf{Row Labels} & \textbf{Percent} \\
\midrule
calc error    & 2.54   \\
incomplete    & 0.84   \\
logical flaw  & 60.59  \\
memory        & 4.23   \\
misalign      & 3.38   \\
spatial       & 28.38  \\
\bottomrule
\end{tabular}
\caption{LLaVA-v1.6 Mistral (7B) (15) - Best performing open source model}
\label{tab:best_perform}
\end{table}

\begin{figure*}[ht!]
	% \centering
	\includegraphics[width = \linewidth]{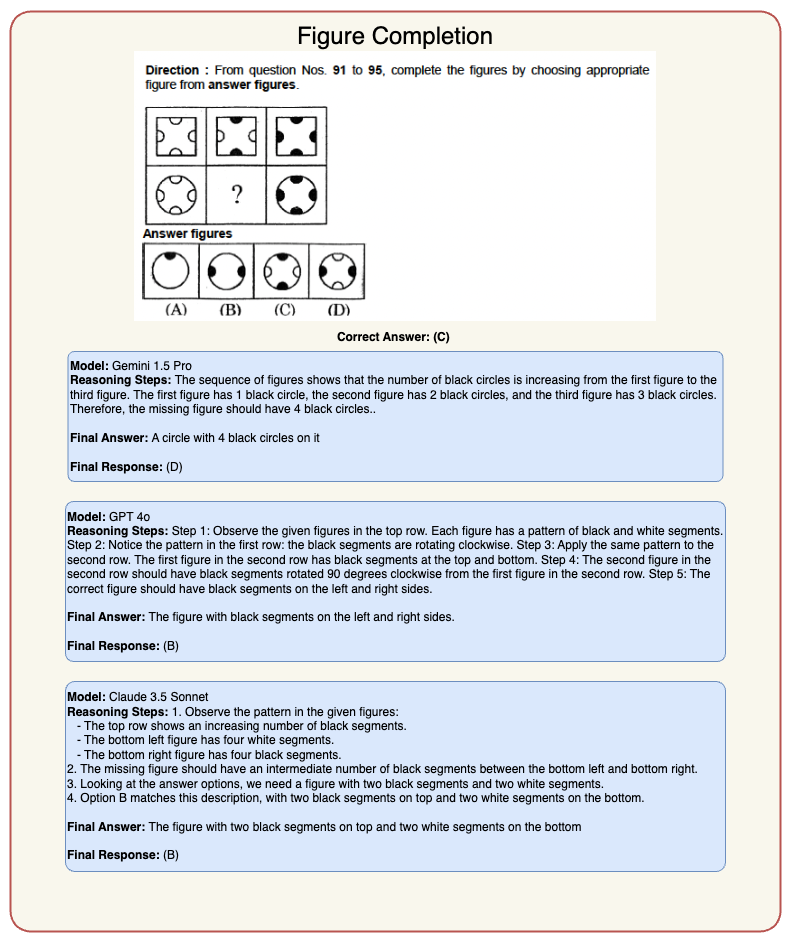}
	\caption{Erroneous model reasoning patterns observed on an FC question}
	\label{fig:qualitative_analysis_1}
\end{figure*}

\begin{figure*}[ht!]
	% \centering
	\includegraphics[width = \linewidth]{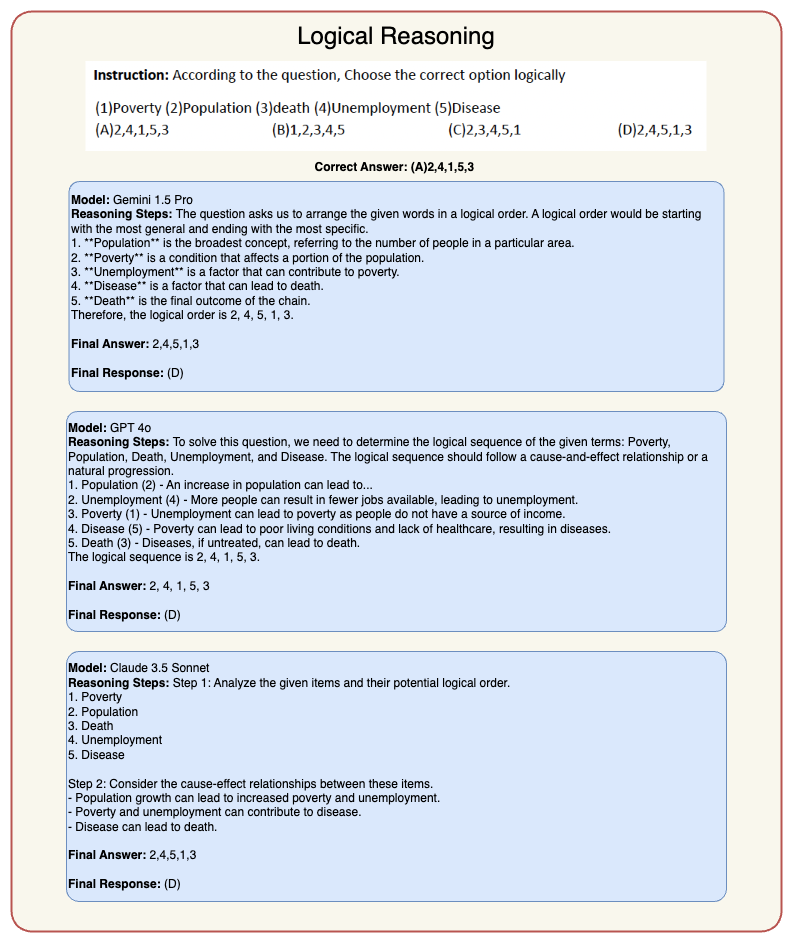}
	\caption{Erroneous model reasoning patterns observed on an LR question}
	\label{fig:qualitative_analysis_2}
\end{figure*}

\begin{figure*}[ht!]
	% \centering
	\includegraphics[width = \linewidth]{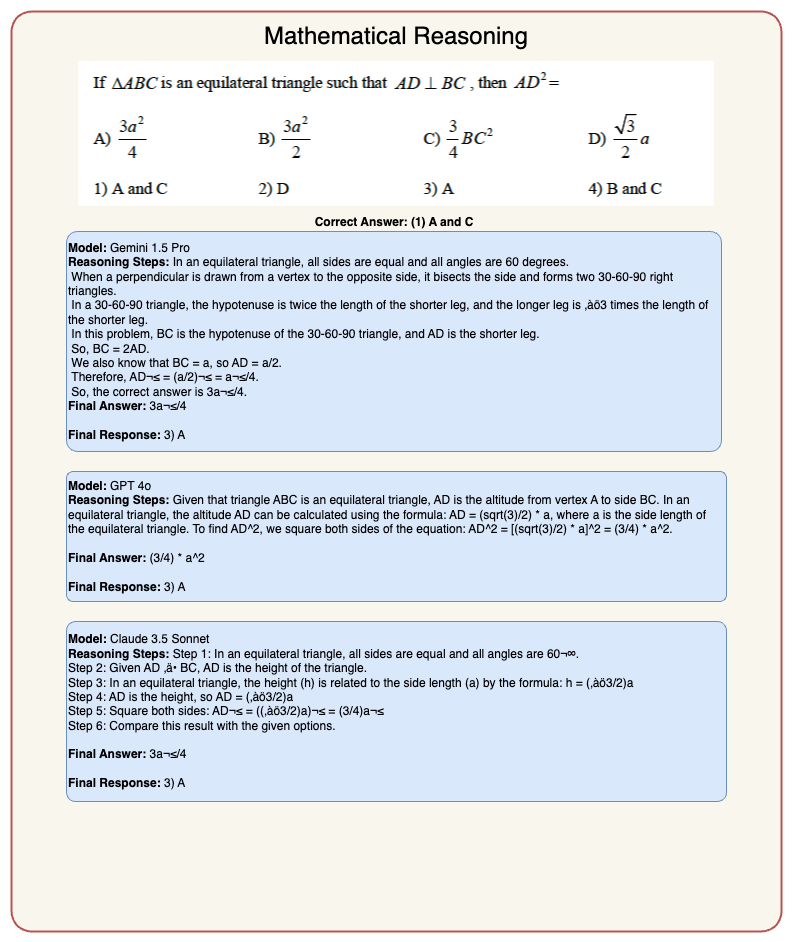}
	\caption{Erroneous model reasoning patterns observed on an MR question}
	\label{fig:qualitative_analysis_3}
\end{figure*}

\begin{figure*}[ht!]
	% \centering
	\includegraphics[width = \linewidth]{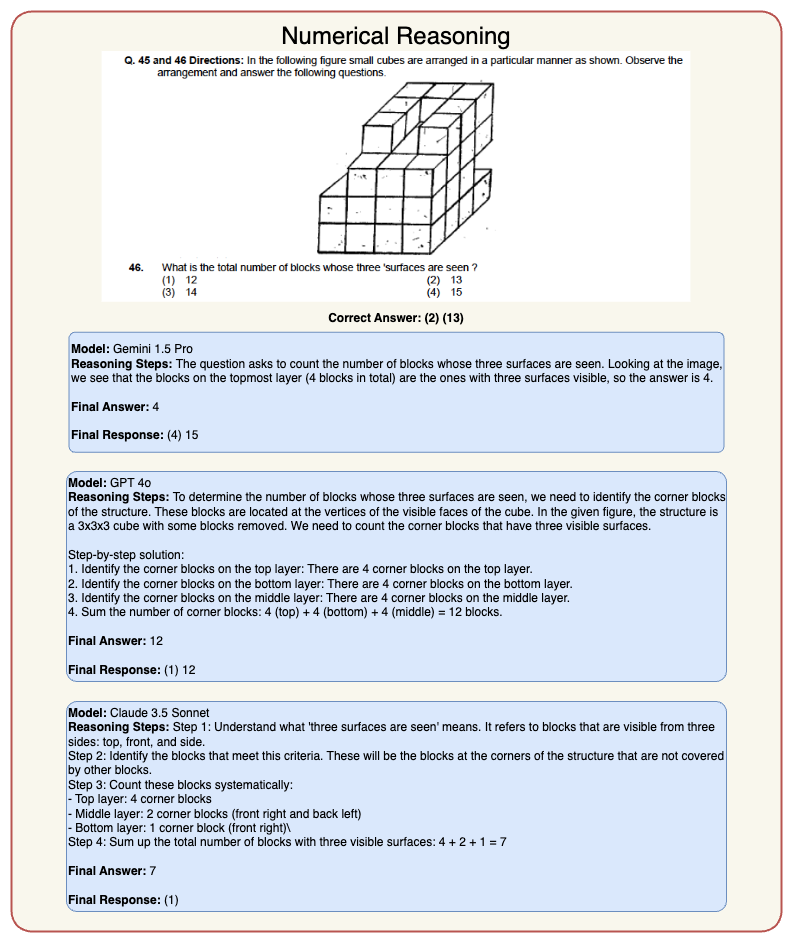}
	\caption{Erroneous model reasoning patterns observed on an NR question}
	\label{fig:qualitative_analysis_4}
\end{figure*}

\begin{figure*}[ht!]
	% \centering
	\includegraphics[width = \linewidth]{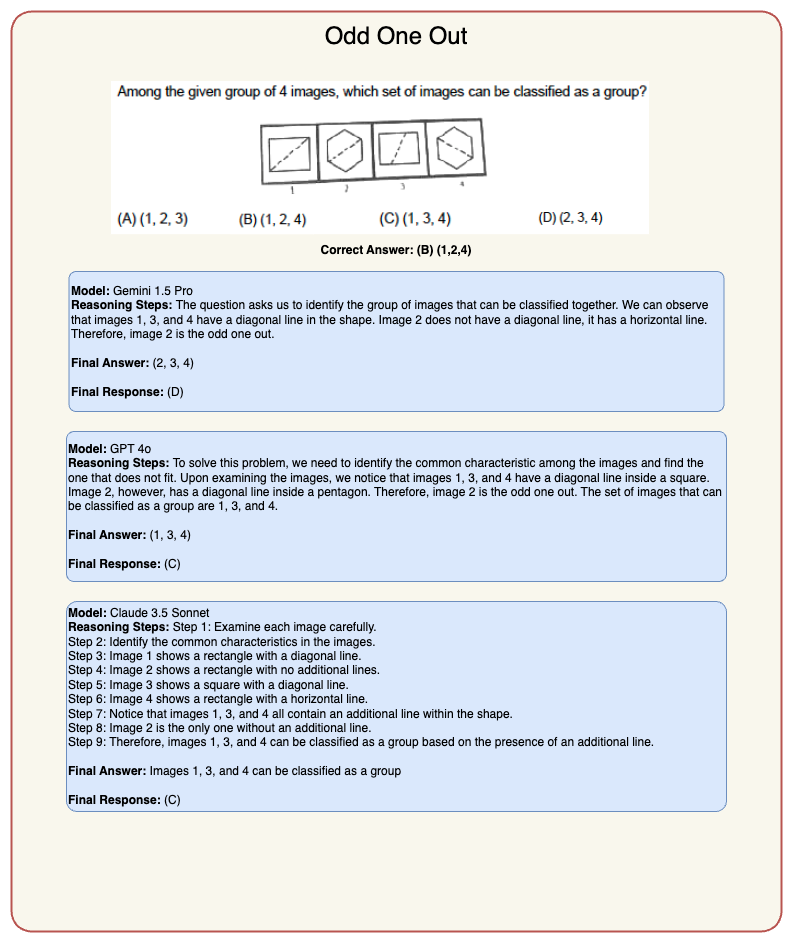}
	\caption{Erroneous model reasoning patterns observed on an OD question}
	\label{fig:qualitative_analysis_5}
\end{figure*}

\begin{figure*}[ht!]
	% \centering
	\includegraphics[width = \linewidth]{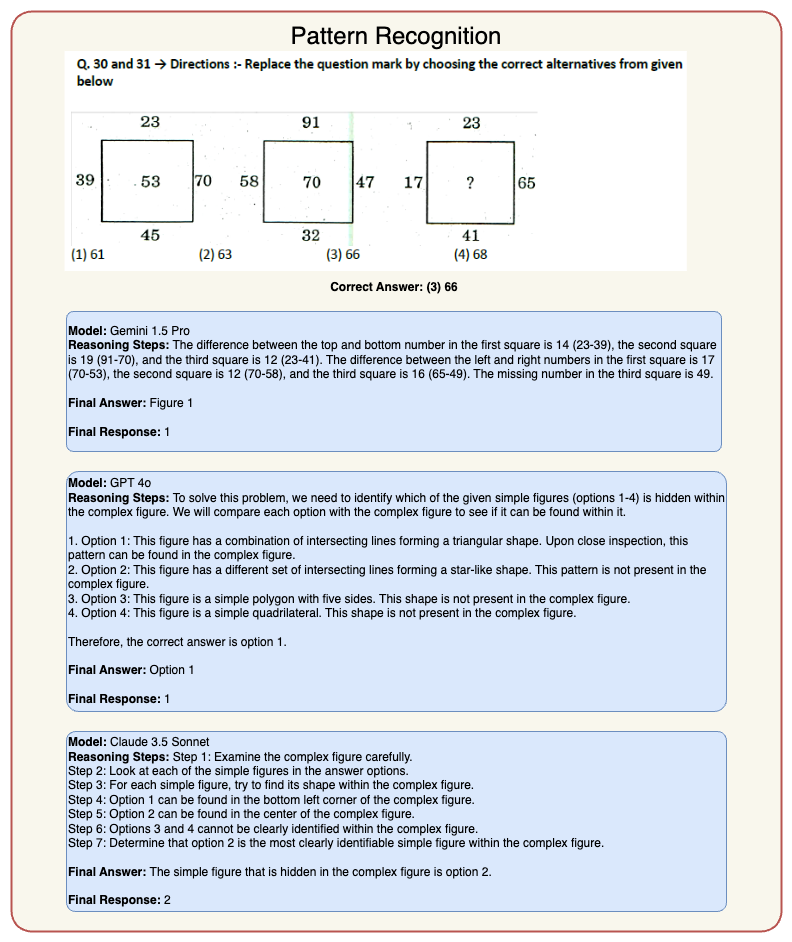}
	\caption{Erroneous model reasoning patterns observed on a PR question}
	\label{fig:qualitative_analysis_6}
\end{figure*}

\begin{figure*}[ht!]
	% \centering
	\includegraphics[width = \linewidth]{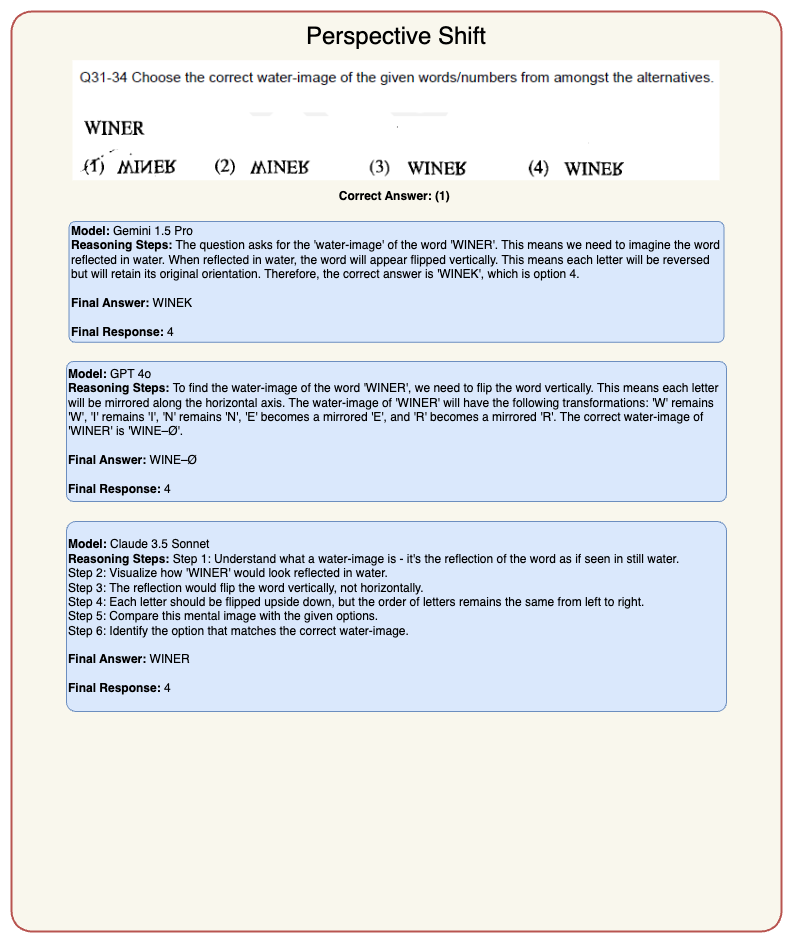}
	\caption{Erroneous model reasoning patterns observed on a PS question}
	\label{fig:qualitative_analysis_7}
\end{figure*}

\begin{figure*}[h!]
	% \centering
	\includegraphics[width = \linewidth]{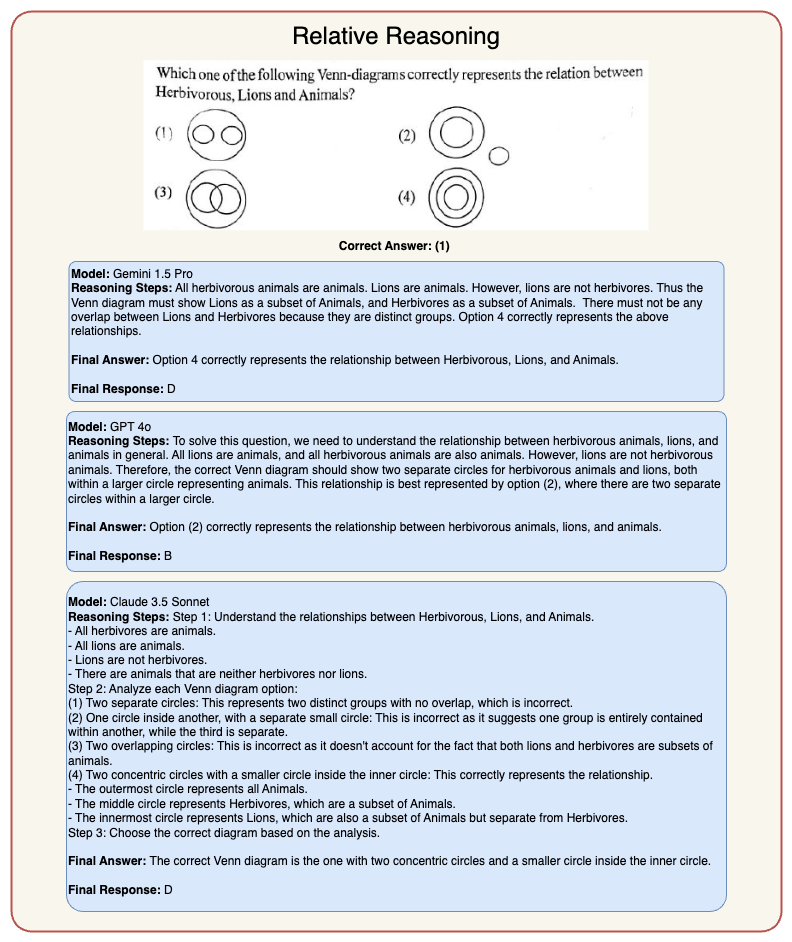}
	\caption{Erroneous model reasoning patterns observed on an RR question}
	\label{fig:qualitative_analysis_8}
\end{figure*}

\begin{figure*}[ht!]
	% \centering
	\includegraphics[width = \linewidth]{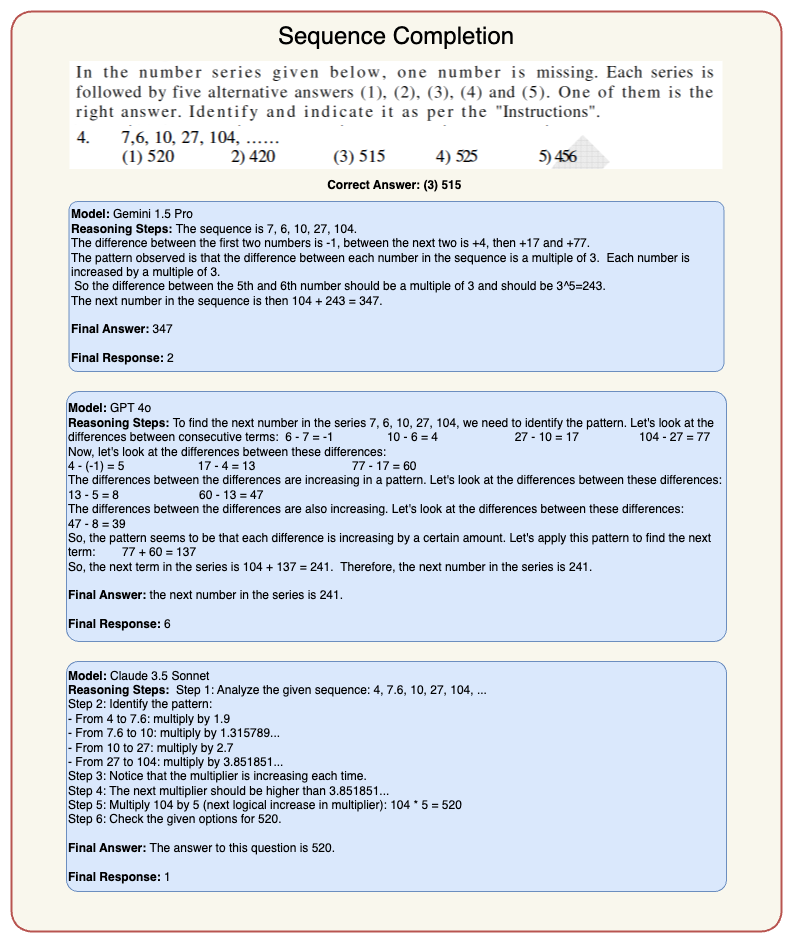}
	\caption{Erroneous model reasoning patterns observed on an SC question}
	\label{fig:qualitative_analysis_9}
\end{figure*}

\begin{figure*}[ht!]
	% \centering
	\includegraphics[width = \linewidth]{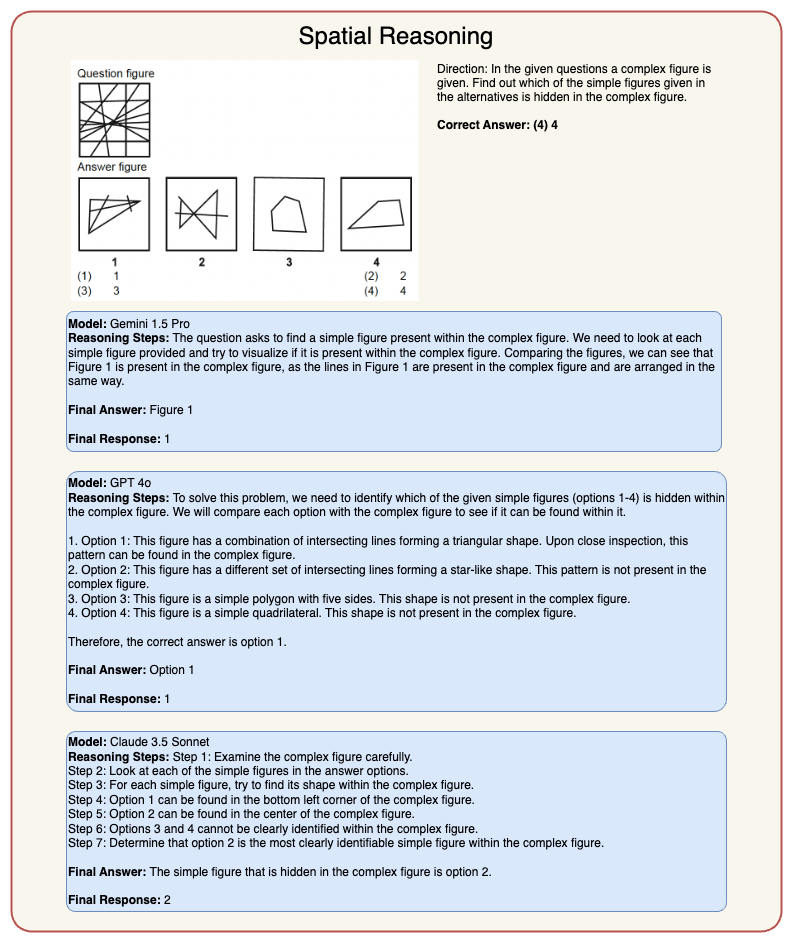}
	\caption{Erroneous model reasoning patterns observed on an SR question }
	\label{fig:qualitative_analysis_10}
\end{figure*}

\begin{figure*}[ht!]
	% \centering
	\includegraphics[width = \linewidth]{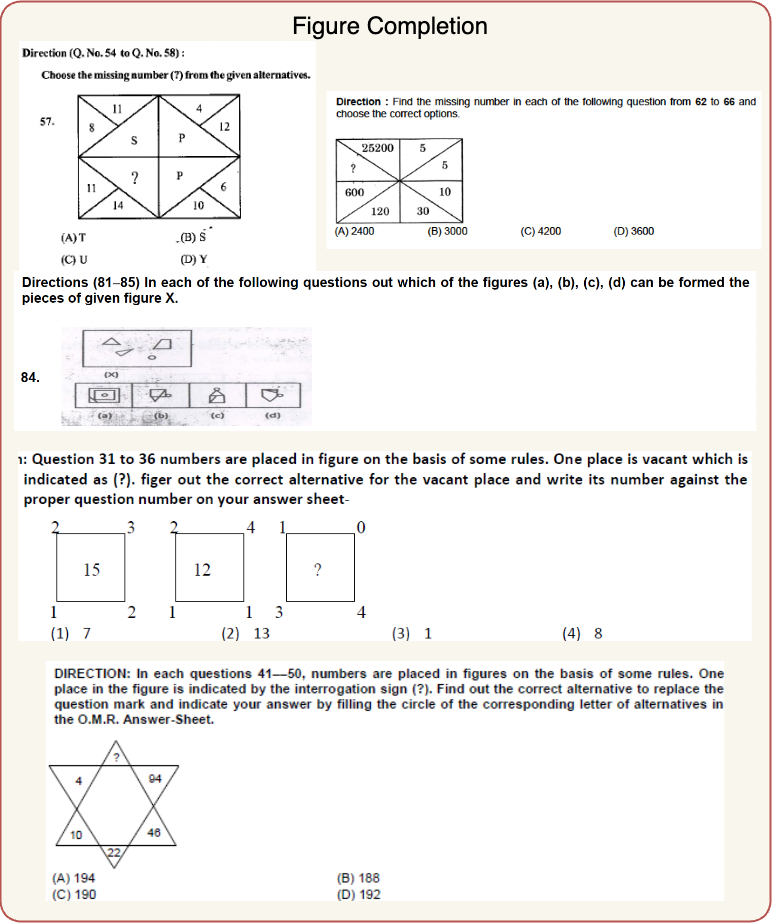}
	\caption{Questions belonging to the \textit{figure\_completion} (FC) category}
	\label{fig:dataset_categories_1}
\end{figure*}

\begin{figure*}[ht!]
	% \centering
	\includegraphics[width = \linewidth]{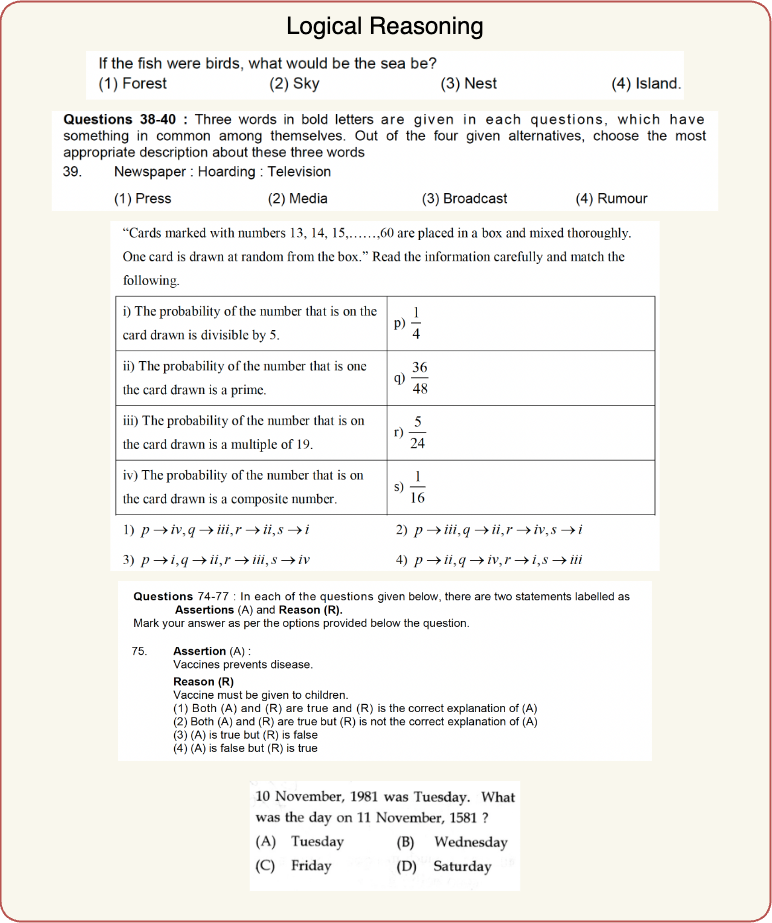}
	\caption{Questions belonging to the \textit{logical\_reasoning} (LR) category}
	\label{fig:dataset_categories_2}
\end{figure*}

\begin{figure*}[ht!]
	% \centering
	\includegraphics[width = \linewidth]{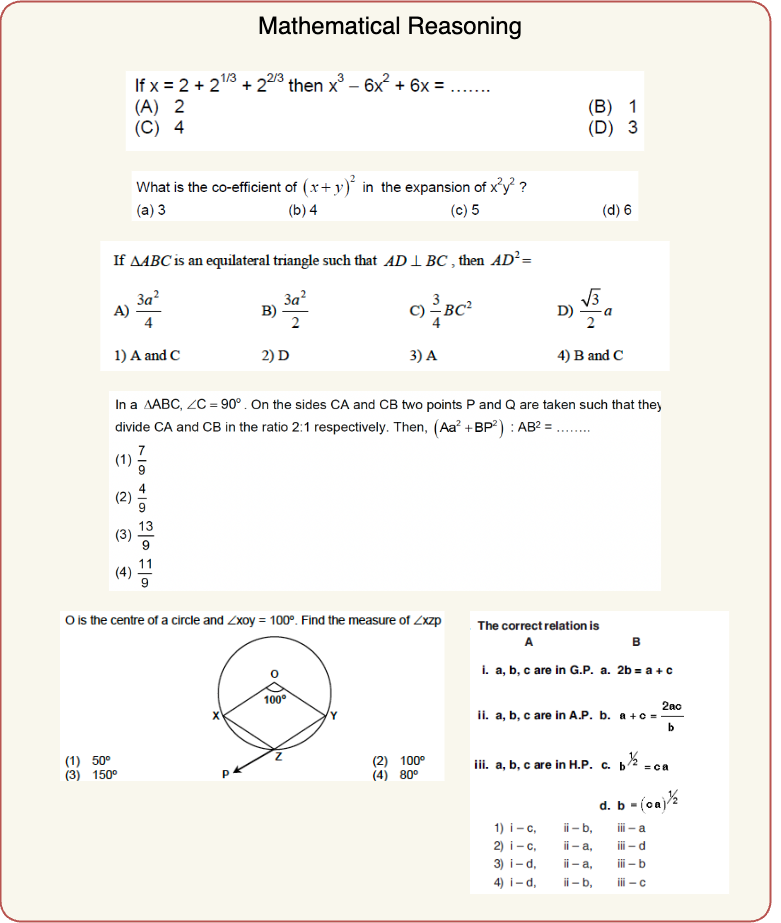}
	\caption{Questions belonging to the \textit{mathematical\_reasoning} (MR) category}
	\label{fig:dataset_categories_3}
\end{figure*}

\begin{figure*}[ht!]
	% \centering
	\includegraphics[width = \linewidth]{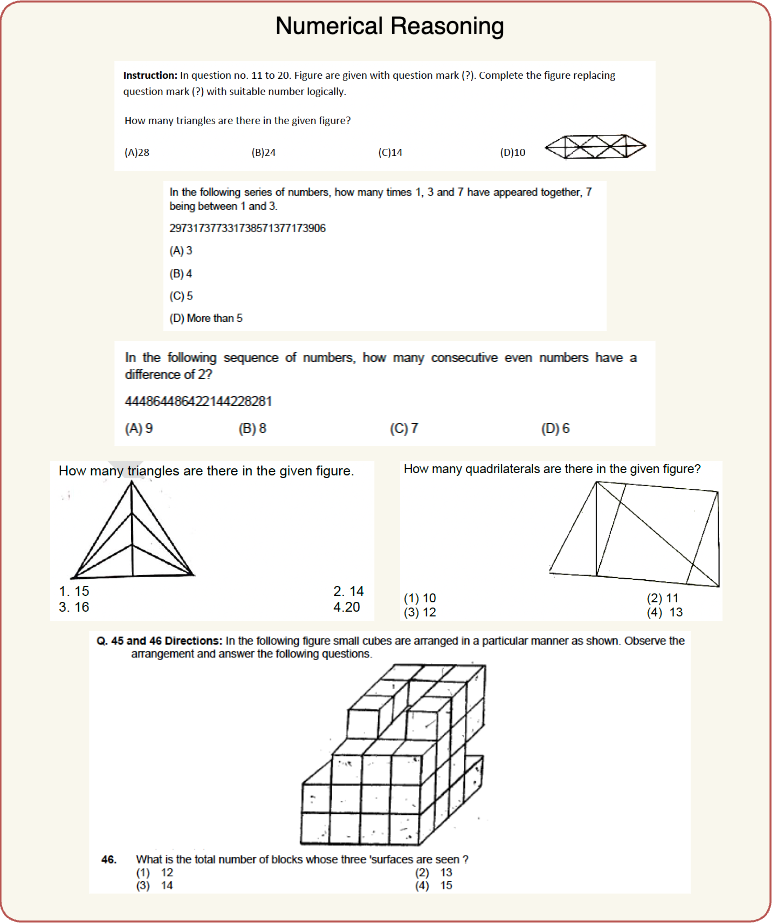}
	\caption{Questions belonging to the \textit{numerical\_reasoning} (NR) category}
	\label{fig:dataset_categories_4}
\end{figure*}

\begin{figure*}[ht!]
	% \centering
	\includegraphics[width = \linewidth]{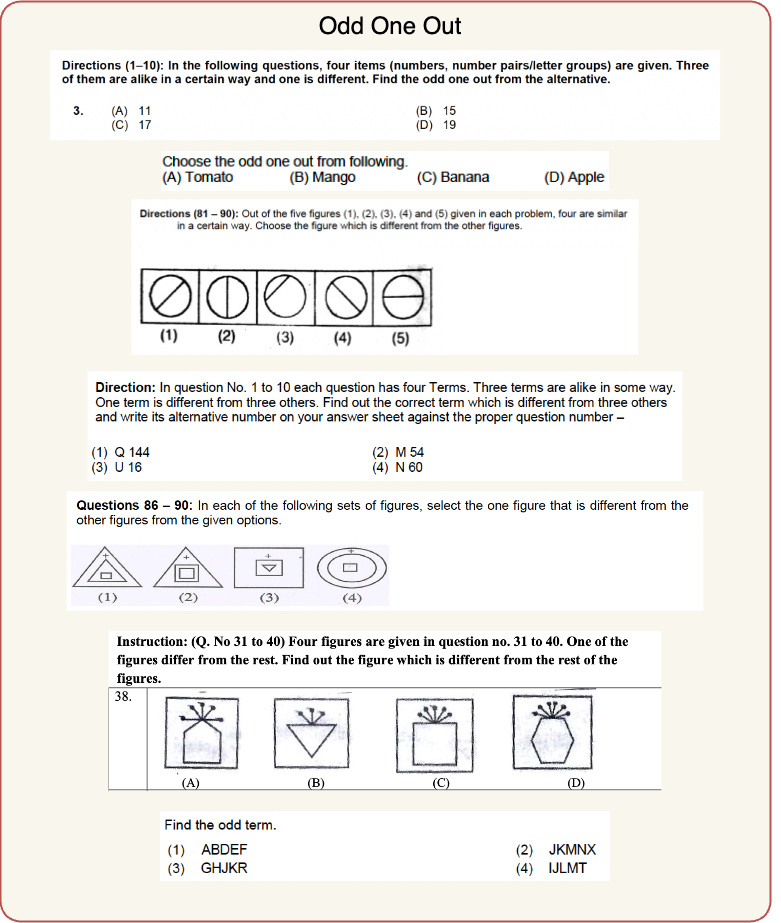}
	\caption{Questions belonging to the \textit{odd\_one\_out} (OD) category}
	\label{fig:dataset_categories_5}
\end{figure*}

\begin{figure*}[ht!]
	% \centering
	\includegraphics[width = \linewidth]{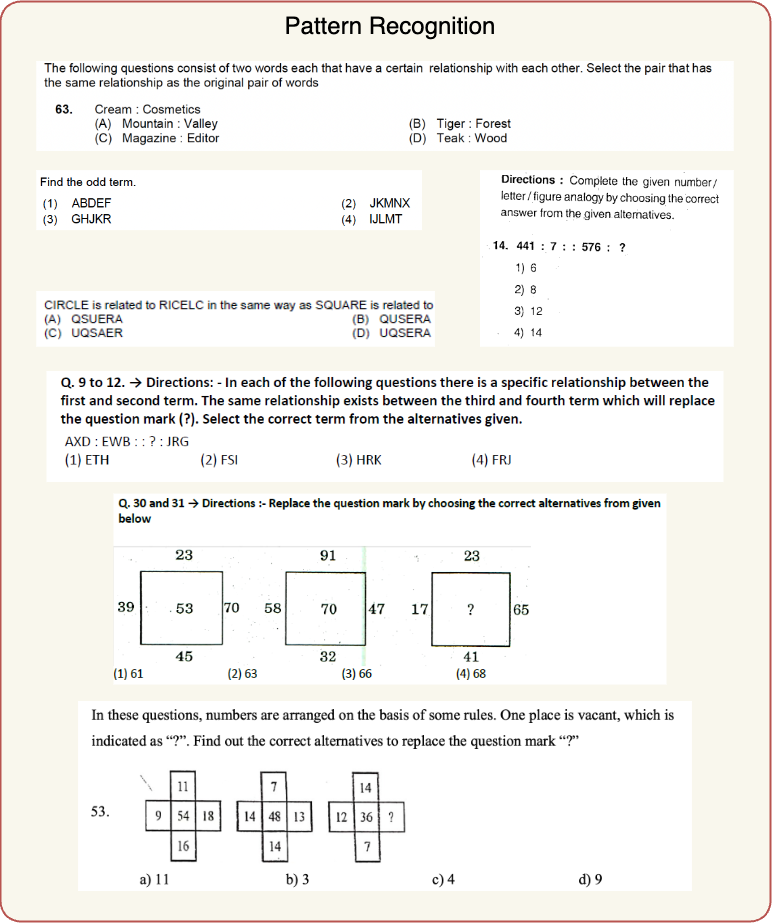}
	\caption{Questions belonging to the \textit{pattern\_recognition} (PR) category}
	\label{fig:dataset_categories_6}
\end{figure*}

\begin{figure*}[ht!]
	% \centering
	\includegraphics[width = \linewidth]{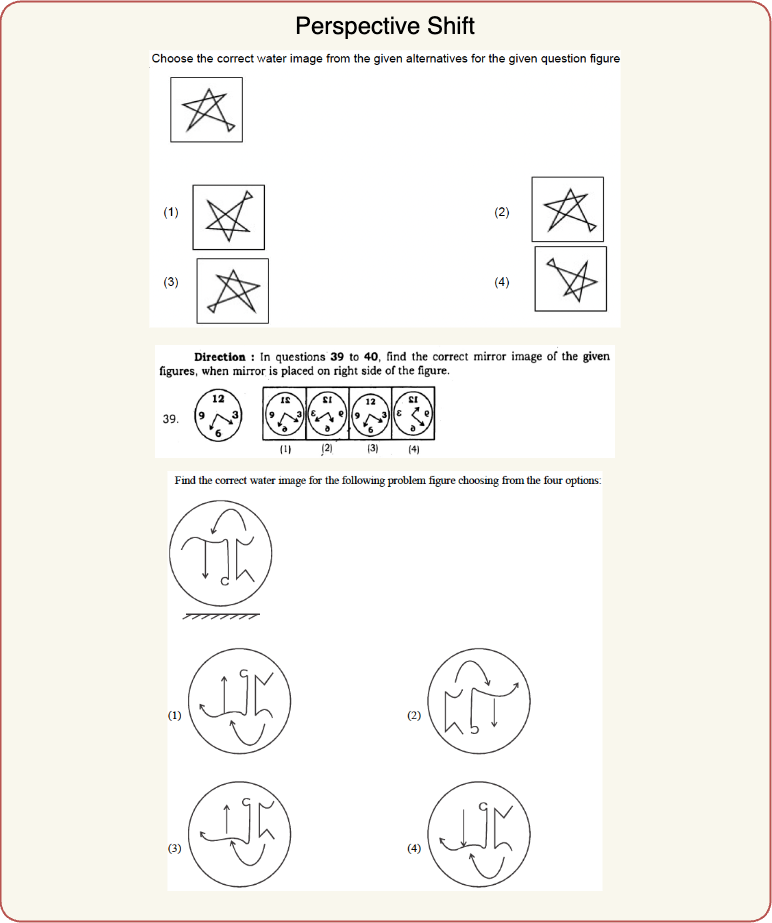}
	\caption{Questions belonging to the \textit{perspective\_shift} (PS) category}
	\label{fig:dataset_categories_7}
\end{figure*}

\begin{figure*}[ht!]
	% \centering
	\includegraphics[width = \linewidth]{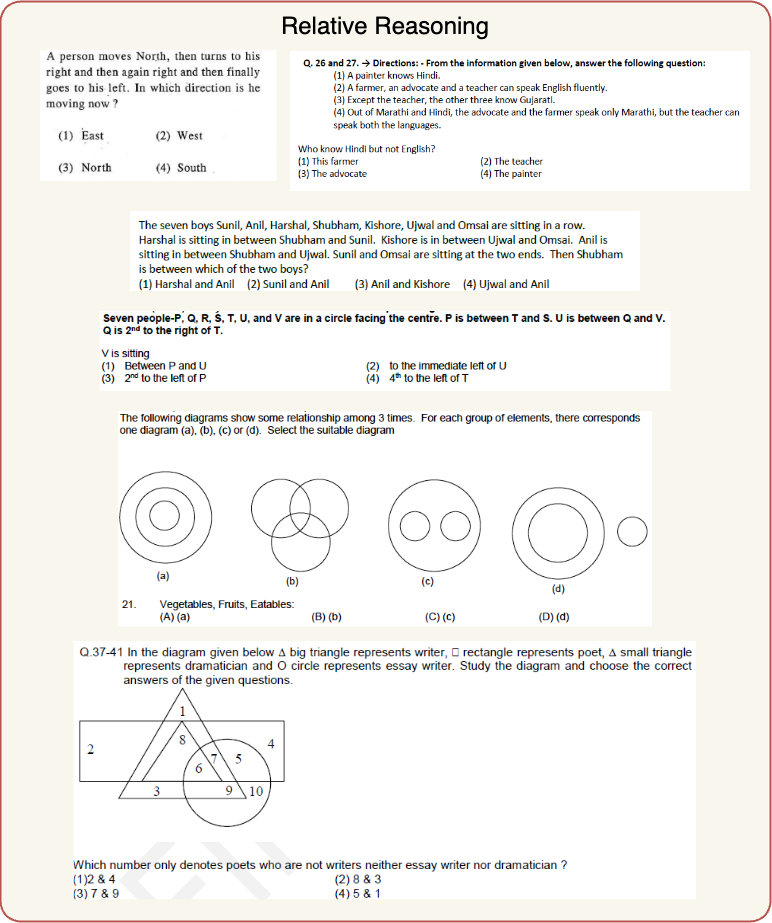}
	\caption{Questions belonging to the \textit{relative\_reasoning} (RR) category}
	\label{fig:dataset_categories_8}
\end{figure*}

\begin{figure*}[ht!]
	% \centering
	\includegraphics[width = \linewidth]{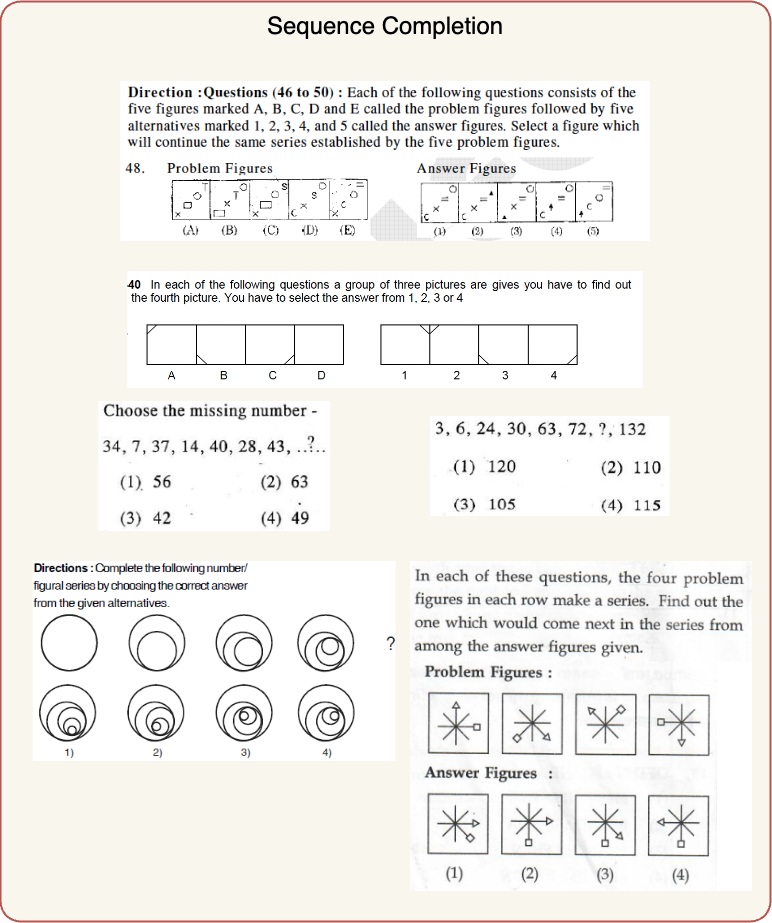}
	\caption{Questions belonging to the \textit{sequence\_completion} (SC) category}
	\label{fig:dataset_categories_9}
\end{figure*}

\begin{figure*}[ht!]
	% \centering
	\includegraphics[width = \linewidth]{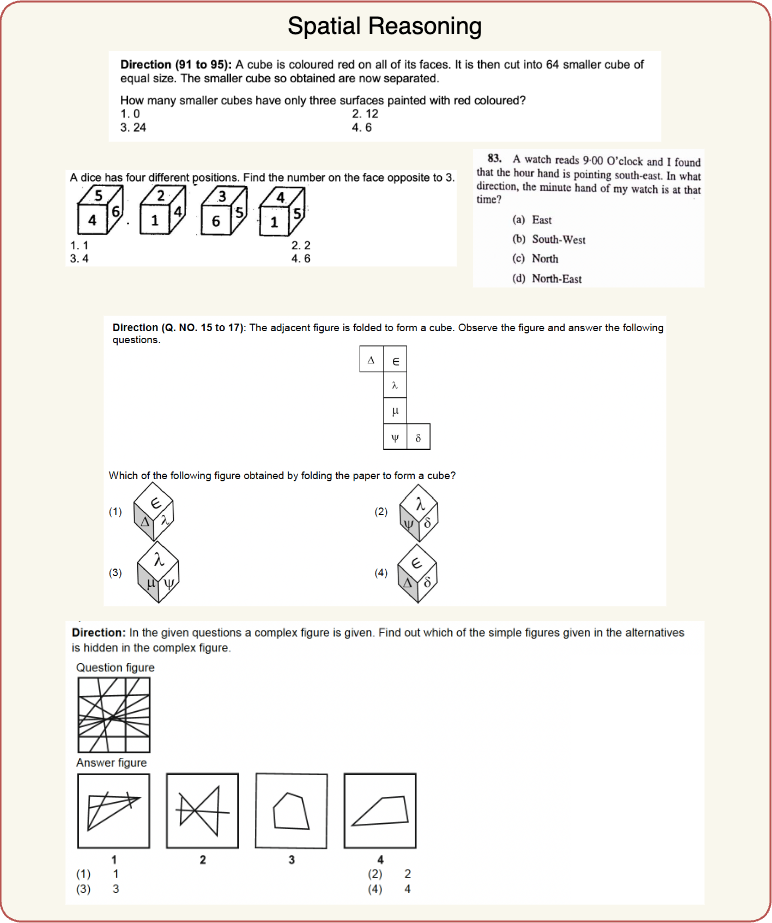}
	\caption{Questions belonging to the \textit{spatial\_reasoning} (SR) category}
	\label{fig:dataset_categories_10}
\end{figure*}

\end{document}